\documentclass{article}

\usepackage{times}
\usepackage{graphicx} 
\usepackage{subfigure} 

\usepackage{natbib}

\usepackage{algorithm}
\usepackage{algorithmic}

\usepackage{hyperref}

\usepackage[accepted]{implicitworkshop2017} 
\usepackage{amsmath,amsthm,mathtools,dsfont}
\usepackage{textcomp}
\usepackage{amssymb}
\usepackage{amsmath}

\icmltitlerunning{Likelihood Estimation for Generative Adversarial Networks}

\begin{document} 

\twocolumn[
\icmltitle{Likelihood Estimation for Generative Adversarial Networks}




\begin{icmlauthorlist}
\icmlauthor{Hamid Eghbal-zadeh}{cp}
\icmlauthor{Gerhard Widmer}{cp}
\end{icmlauthorlist}
\icmlaffiliation{cp}{Department of Computational Perception, Johannes Kepler University of Linz, Austria}

\icmlcorrespondingauthor{Hamid Eghbal-zadeh}{hamid.eghbal-zadeh@jku.at}

\icmlkeywords{generative adversarial networks, likelihood estimation, Adversarial networks, generative models}

\vskip 0.3in
]



\printAffiliationsAndNotice{}  

\begin{abstract} 
We present a simple method for assessing the quality of generated images in Generative Adversarial Networks (GANs). The method can be applied in any kind of GAN without interfering with the learning procedure or affecting the learning objective.
The central idea is to define a likelihood function that correlates with the quality of the generated images.
In particular, we derive a Gaussian likelihood function from the distribution of the embeddings (hidden  activations) of the real images in the discriminator, and based on this, define two simple measures of how likely it is that the embeddings of generated images are from the distribution of the embeddings of the real images.
This yields a simple measure of fitness for generated images, for all varieties of GANs.
Empirical results on CIFAR-10 demonstrate a strong correlation between the proposed measures and the perceived quality of the generated images.
\end{abstract}

\section{Introduction}
\label{intro}
Generative Adversarial Networks (GANs) \citep{goodfellow2014generative} are one of the big discoveries in deep learning in recent years.
GANs can learn to generate realistic images through an iterative game between a generator network and a discriminator network.
Although GANs are able to learn the Probability Density Function (PDF) underlying a set of real images, they can not explicitly provide a likelihood for (i.e., effectively evaluate) this estimated PDF. In other words, they cannot provide a direct estimate of the probability of an image to come from the true underlying distribution.
As an alternative, many different variations of GANs have been proposed where some other measure of goodness is provided to assess the quality of the generated images.
For instance, the Wasserstein GAN \citep{arjovsky2017wasserstein} relies on the \textit{Wasserstein distance} between the activations of the real and the generated images in the discriminator.
Other approaches have been developed that not only use a measure of distance in the discriminator, but also consider a \textit{probability measure} in calculating distances.
In the Cramer-GAN \citep{bellemare2017cramer}, the authors use a distance from the family of integral probability metrics (IPM) which measures the distance between probability distributions of generated and real images in the discriminator's activations.
As another example, MMD-GAN \citep{li2017mmd} relies on the maximum mean discrepancy (MMD) between two distributions of the real and fake images.

While all these models rely on a measure of distance, they all do so by using their specific distances as an objective for training the GAN.
Hence, they can not be ported into other kinds of GANs with different training objectives.
In \citep{salimans2016improved}, the \emph{inception score} is introduced for comparing the quality of the generated images across different Generative models.
This is basically the Inception Model \citep{szegedy2016rethinking} trained on ImageNet to get both the conditional and marginal label distribution for the generated image.
Inception score requires lots of samples (as suggested, $50k$) and also requires a model trained on ImageNet.

In this paper, we propose \textit{Likelihood Estimation for Generative Adversarial Networks (LeGAN)}, a general measure of goodness, based on a likelihood function defined for the discriminator, which can be applied to any kind of GAN.
LeGAN does not interfere with the training of GANs and does not need any pre-training.
We will attempt to show the correlation between this measure and perceived image quality in an experiment with CIFAR-10 data.

\section{Generative Adversarial Networks}
\label{gans}

In GANs, a discriminator $D$ tries to distinguish between real and generated (fake) images while competing with a generator $G$ that tries to fool the discriminator $D$ by generating realistic images.
The loss of the discriminator $D$ in the vanilla GAN \citep{goodfellow2014generative} is defined as:
\begin{equation}
\label{eq:disc_loss}
\mathcal{L}_{D} = \mathbb{E}_{\boldsymbol{x} \sim P_{\mathit{real}} } \big[ \log D(\boldsymbol{x}) \big] + \mathbb{E}_{\boldsymbol{z}\sim P_{\mathit{fake}} } \big[ \log (1 - D(G(\boldsymbol{z}))) \big] 
\end{equation}

(where it is assumed that for a given input image (vector) $\boldsymbol{x}$, the discriminator $D$ outputs a number -- the estimated probability of the image coming from the real set); and the loss of the generator $G$ is defined as:
\begin{equation}\label{eq:gen_loss}
\mathcal{L}_{G}=\mathbb{E}_{\boldsymbol{z} \sim P_{\mathit{fake}} } \big[ -\log D(G(\boldsymbol{z})) \big] 
\end{equation}

where
$\boldsymbol{z}$ is a random observation from a distribution $Z$; generator $G$ creates a fake image using this $\boldsymbol{z}$.

A modified version of the losses defined above using a least squares criterion is introduced in Least Squares GAN (LSGAN) \citep{mao2016least} as follows:
\begin{equation}\label{eq:ls_disc_loss}
{\mathcal{L}^{\mathit{ls}}_{D}}=\frac{1}{2}\mathbb{E}_{\boldsymbol{x} \sim P_{\mathit{real}} } \big[ (D(\boldsymbol{x})-1)^2 \big] + \frac{1}{2}\mathbb{E}_{\boldsymbol{z}\sim P_{\mathit{fake}} } \big[D(G(\boldsymbol{z}))^2 \big] 
\end{equation}

and the loss of the generator $G$ in LSGAN is defined as:
\begin{equation}\label{eq:ls_gen_loss}
{\mathcal{L}^{\mathit{ls}}_{G}}=\frac{1}{2}\mathbb{E}_{\boldsymbol{z} \sim P_{\mathit{fake}}} \big[ ( D(G(\boldsymbol{z}))-1)^2 \big] 
\end{equation}

None of the losses defined above provide or use a measure that correlates with the quality of the generated images while the training procedure of GANs.
In contrast, in the Wasserstein GAN the generator attempts to minimize the Wasserstein distance between $D(x)$ and $D(G(z))$.
As this distance approaches zero, the two distributions become more and more similar as the quality of the generated images improves.

In the following section we explain LeGAN, a tool that can be used with any GAN to provide a meaningful measurement that correlates with the quality of the generated images.


\section{Likelihood Estimation for Generative Adversarial Networks}
\label{legan}

\begin{figure}[t]
\vskip 0.2in
\begin{center}
\centerline{\includegraphics[width=\columnwidth]{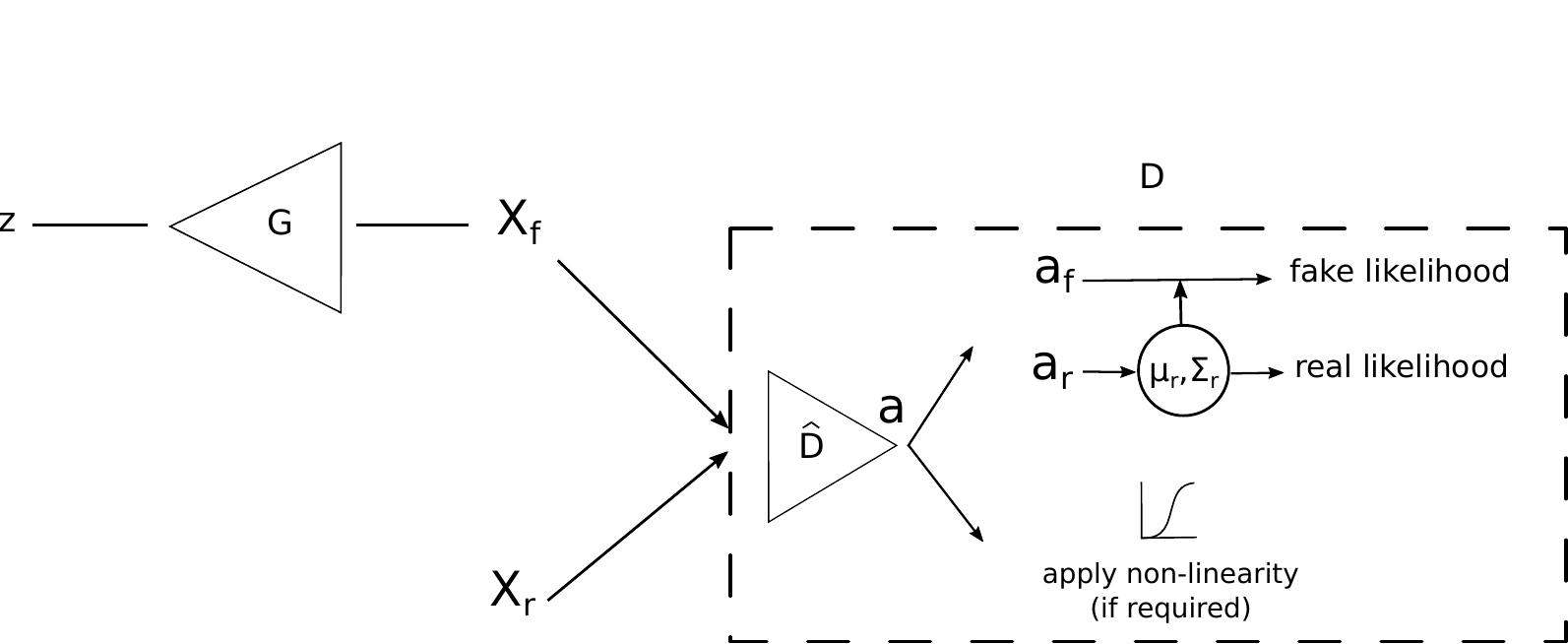}}
\caption{Block diagram of LeGAN. $G$: the generator. $\hat{D}$: the discriminator without the final nonlinearity, creating embeddings. $D$: the discriminator. $X_r$: real images. $X_f$: fake (generated) images in one batch of data. $\mu_r$ and $\sigma^2_r$: mean and variance of the embeddings of real images. $a_r$: embeddings of real images. $a_f$: embeddings of fake images. $z$: random vector from a known distribution.}
\label{fig:block_diag}
\end{center}
\vskip -0.2in
\end{figure}

In this section, we describe the proposed measurement to assess the quality of generated images in GANs.
The underlying idea is very simple: by estimating the distribution of the embeddings of the real images in the discriminator, we define a likelihood function. This function can then measure the likelihood of any arbitrary image based on the image's discriminator embedding.
For the real images this likelihood is expected to be higher than for fake images. 
When the generated images are no longer distinguishable from the real ones, the likelihood for the generated images should also become similar to the likelihood of real images.
Using central limit theorem \citep{gnedenko1954limit}, we choose to model the embeddings of the discriminator using a Gaussian distribution with the assumption that our discriminator embeddings will be roughly normally distributed (which actually turns out to be the case in the experiments mentioned in the next section).
But this can be replaced with any other distribution.

A block diagram of LeGAN is provided in Figure \ref{fig:block_diag}.
The generator $G$ generates (fake) images $X_f$, given random vectors $z$ from a known distribution as input.
The discriminator $D$, trained with $X_f$ as negative examples and real images $X_r$ as positive examples, tries to distinguish between real and fake samples.
In $D$, $a$ is the embedding\footnote{These embeddings are the hidden activations of the last layer in the discriminator network before applying the final activation function. Since we want the distribution of the embeddings and not a probability, applying the sigmoid non-linearity is not required.} representation of images which then can be represented as \emph{probability} by applying activation functions such as \emph{sigmoid}. Hence, $a_r$ represents the embeddings of real images and $a_f$ represents the embeddings of the fake images created by the discriminator network.

Now we define the following likelihood for a given image embedding $a =\hat{D}(\boldsymbol{x})$ with an assumption of $a$ being Gaussian distributed:
\begin{equation}\label{eq:mdn_bottleneck_loss} 
\ell_\mathcal{G}(a) = P(a \mid \mathcal{G})
 = \mathcal{N} \Big( a; \mu_{r}, \sigma^2_{r} \Big)
\end{equation}
where $\hat{D}$ is the discriminator $D$ without its final activation function.
The Gaussian $\mathcal{G}$ is defined by the parameters $\mu_r$ (mean) and $\sigma^2_r$ (variance) learned from the embeddings of real images in  $\hat{D}$.

Based on the likelihood function $\ell_\mathcal{G}(a)$ introduced above, we propose two measurements for GANs:
\begin{equation}\label{eq:likelihood_diff}
\ell_\mathit{diff} = \ell_\mathcal{G}(\bar{a}_r) - \ell_\mathcal{G}(\bar{a}_f)
\end{equation}

as the likelihood difference and
\begin{equation}\label{eq:likelihood_ratio}
\ell_\mathit{ratio} = \frac{\min(\ell_\mathcal{G}(\bar{a}_r) , \ell_\mathcal{G}(\bar{a}_f))} {\max(\ell_\mathcal{G}(\bar{a}_r) , \ell_\mathcal{G}(\bar{a}_f))}
\end{equation}
as the likelihood ratio, where $\bar{a}_r$ and $\bar{a}_f$ are the average of the embeddings in a batch for real and fake images, respectively.

Both of these measures relate to how well the embeddings of fake images match to the Gaussian likelihood function parameterized by the embeddings of real images. When the generated images are close enough to the real images, the distribution of $a_r$ becomes more similar to $a_f$ and therefore $a_f$ fits better in $\ell_\mathcal{G}$ and achieves similar likelihoods as $\ell_\mathcal{G}(a_r)$.

\section{Distance correlation between images and discriminator embeddings}
\label{sec:weight_clipping}

LeGAN provides a measure based on the discriminator's embedding distribution of real and fake images which correlates with the quality of the generated images.
For this to work, we would need to find a way to relate the distance measure in the images to the distance measure between the embeddings of the discriminator.
We achieve this property by making our discriminator to be 1-Lipschitz continuous.
Assuming $x$ and $y$ two arbitrary images, a discriminator $\hat{D}$ is 1-Lipschitz continuous if:

\begin{equation}\label{eq:lipschitz}
d_\textbf{A}(\hat{D}(x),\hat{D}(y)) \leq d_\textbf{X}(x,y)
\end{equation}
where $d_\textbf{A}$ is a distance defined on $\textbf{A}$ which is the embedding space of $\hat{D}$ and a distance $d_X$ defined on the images space $\textbf{X}$.

To ensure that the embeddings of $\hat{D}(x)$ are bounded, we apply weight-clipping similar to what was proposed in \citep{arjovsky2017wasserstein} to satisfy the 1-Lipschitz continuity of $\hat{D}(x)$ by restricting our weights in the range of $[-0.02,0.02]$.
Since the weights are bounded in this range which is smaller than one, therefore the output of the networks which is sum of the products of the inputs with the weights, is also bounded.
Besides, to enforce the network to learn the weights with smaller values, we apply the \emph{L2 norm} penalty on the weights of the discriminator.

\begin{figure}
\vskip 0.2in
\centering     
\subfigure[Without weight clipping \newline at epoch 1.]{\label{fig:a}\includegraphics[width=0.49\columnwidth]{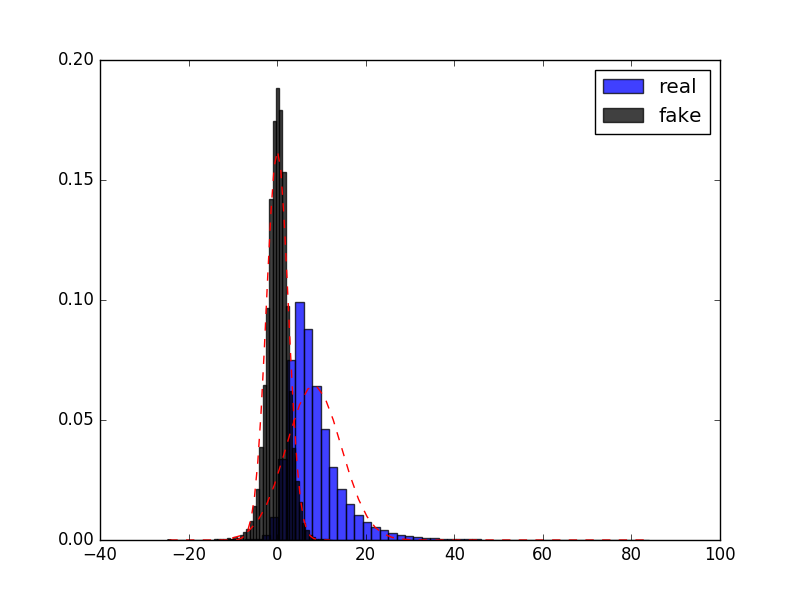}}
\hfill
\subfigure[With weight clipping at epoch 1.]{\label{fig:b}\includegraphics[width=0.49\columnwidth]{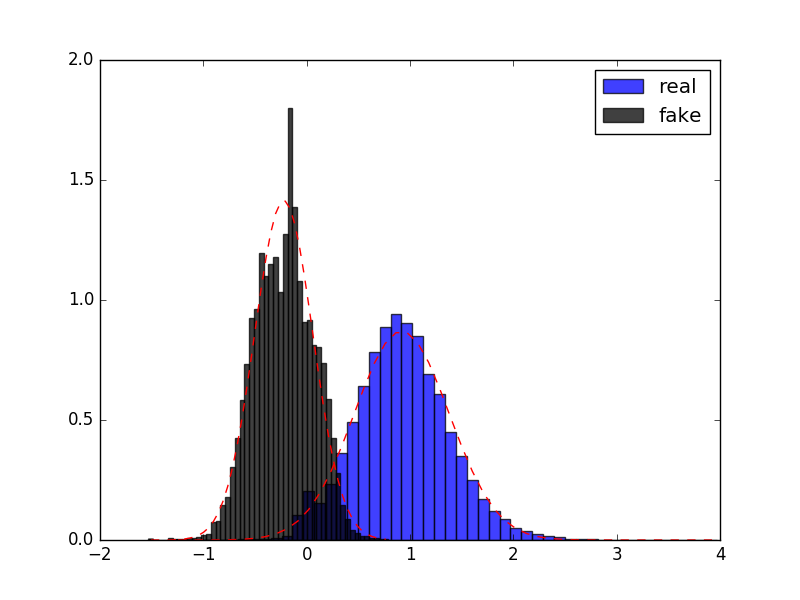}}
 \vskip\baselineskip
\subfigure[Without weight clipping \newline at epoch 13.]{\label{fig:a}\includegraphics[width=0.49\columnwidth]{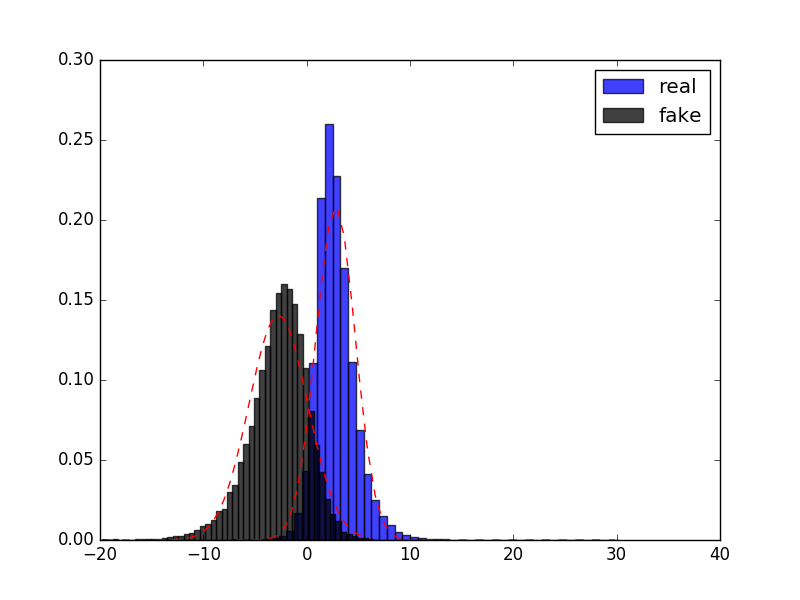}}
\hfill
\subfigure[With weight clipping at epoch 13.]{\label{fig:b}\includegraphics[width=0.49\columnwidth]{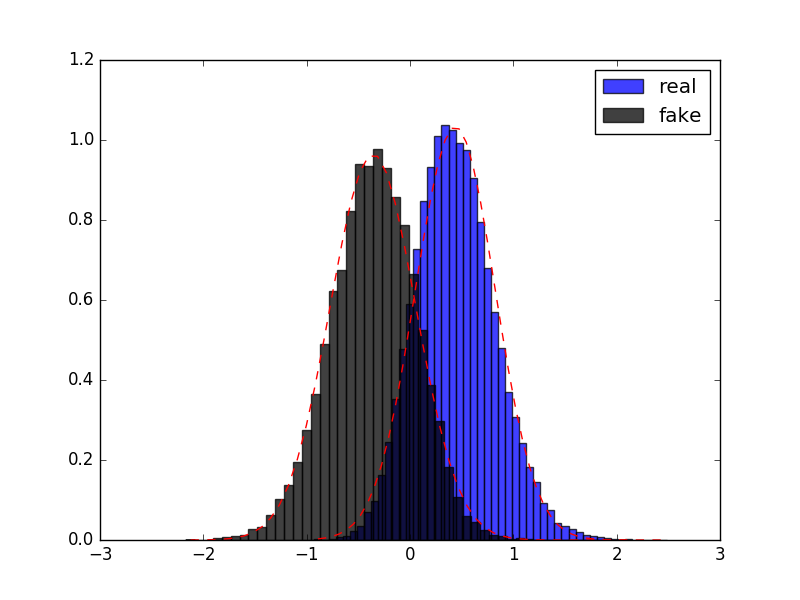}}
\vskip\baselineskip
\subfigure[Generated images at epoch 1.]{\label{fig:a}\includegraphics[width=0.49\columnwidth]{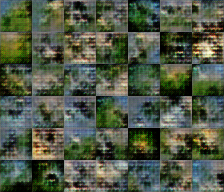}}
\hfill
\subfigure[Generated images at epoch 13.]{\label{fig:b}\includegraphics[width=0.49\columnwidth]{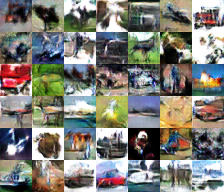}}
\caption{First and second row: Histograms of real and fake embeddings in $\hat{D}(x)$ of the vanilla GAN \citep{goodfellow2014generative}. The third row shows samples of generated images at epoch one and also at epoch $13$. Please note the very different ranges in the $x$ and $y$  axis. Third row: generated images.}
\vskip -0.2in
\label{fig:activation_dist}
\end{figure}

Figure \ref{fig:activation_dist} demonstrates the effect of the weight-clipping on the distribution of the discriminator.
As can be seen in Figure \ref{fig:activation_dist}.a and Figure \ref{fig:activation_dist}.c, the distribution of the embeddings does not become more similar as the model trains and generates better images.
In contrast, it can be seen in Figure \ref{fig:activation_dist}.b and Figure \ref{fig:activation_dist}.d that at first the real and fake embedding distributions are far apart and as the model trains and generates better images, the distribution of real and fake embeddings becomes more similar and the means of the two clusters become closer.
More related examples can be found in the Appendix. 

\section{Empirical results}
\label{empirical_results}
\begin{figure}[ht]
\vskip 0.2in
\begin{center}
\centerline{\includegraphics[width=\columnwidth]{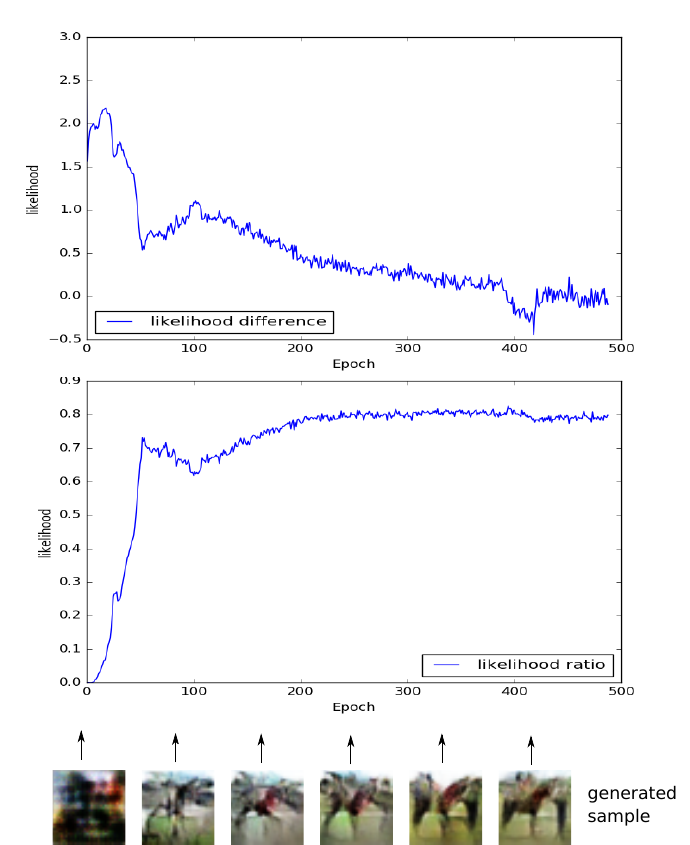}}
\caption{The proposed measures with generated samples using a vanilla GAN\citep{goodfellow2014generative}.}
\label{fig:org_gan_samples}
\end{center}
\vskip -0.2in
\end{figure}

\begin{figure}[ht]
\vskip 0.2in
\begin{center}
\centerline{\includegraphics[width=\columnwidth]{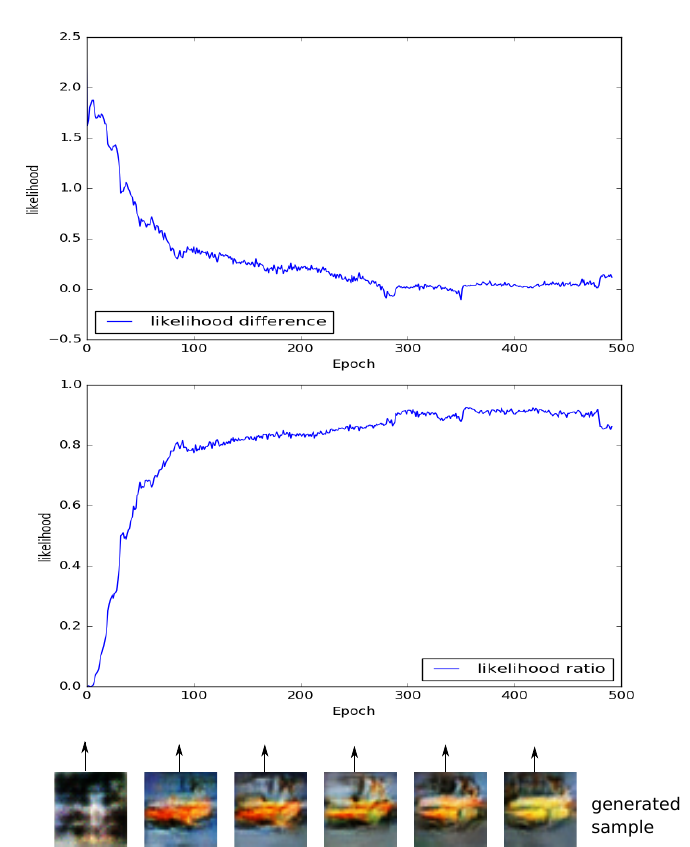}}
\caption{The proposed measures with generated samples using an LSGAN \citep{mao2016least}.}
\label{fig:ls_gan_samples}
\end{center}
\vskip -0.2in
\end{figure}

In this section, you can find the empirical results on the similarity measures of LeGAN.
Figure \ref{fig:org_gan_samples} shows how the quality of generated images is correlated with the LeGAN measures for an vanilla GAN.
Also in Figure \ref{fig:ls_gan_samples} another example is provided for an LSGAN to demonstrate this correlation.
As can be seen in both example, while the quality of generated images increases with more training updates, $\ell_\mathit{diff}$ decreases which shows that the difference in the likelihood of the real and generated images decreases.
By looking at $\ell_\mathit{ratio}$ we can also observe that the ratio between the likelihood of real and fake images increases.
For a better understanding of the behavior of LeGAN, we provided more experimental results in the appendix.

\section{Conclusion}
\label{conclusion}
In this paper, we proposed two measures for the fitness of the generated images based on a Gaussian likelihood function from the distribution of the embeddings of the real images from the discriminator.
We provided experimental results CIFAR-10 demonstrating that our measure correlates with the quality of the images generated with any GANs.
We used our measurement with the vanilla GAN and LSGAN which by default do not have a measure of fitness.
We also compared our measures with the Wasserstein distance in Wasserstein GAN \footnote{These results are provided in the appendix.}, showing our measures show correlation with the Wasserstein distance.

\section*{Acknowledgments} 
 
This work was supported by the Austrian Ministries BMVIT and BMWFW, and the Province of Upper Austria, via the COMET Center SCCH.
We also gratefully acknowledge the support of NVIDIA Corporation with the donation of a Titan X GPU used for this research.


\bibliography{refs}
\bibliographystyle{icml2017}

\clearpage
\onecolumn

\section{Appendix}
\subsection{Extended Empirical Results}
\label{extended_results}

\begin{figure}
\vskip 0.2in
\centering     
\subfigure[${\mathcal{L}_{D}}$ (Wasserstein distance) and ${\mathcal{L}_{G}}$.]{\label{fig:ap:vanilla_gan_loss}\includegraphics[width=0.24\columnwidth]{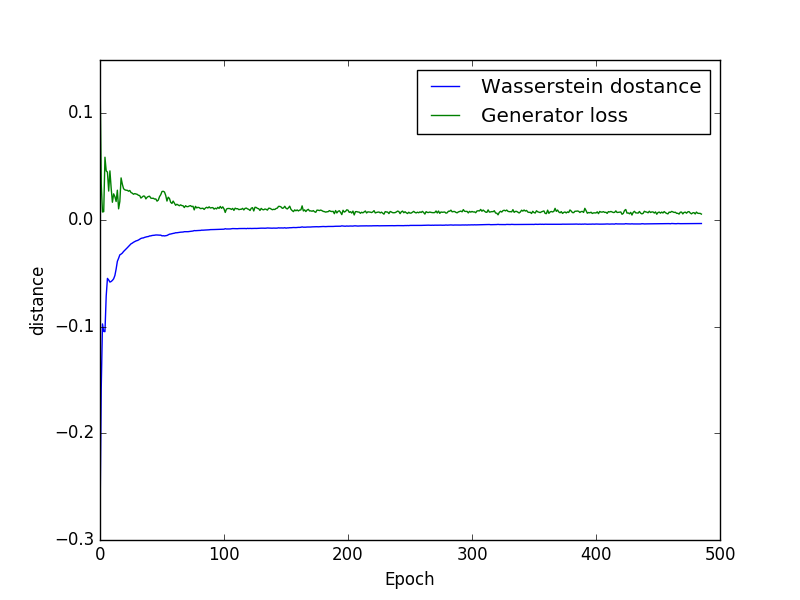}}
\subfigure[$\ell_\mathit{real}$ and $\ell_\mathit{fake}$.]{\label{fig:ap:vanilla_gan_lks}\includegraphics[width=0.24\columnwidth]{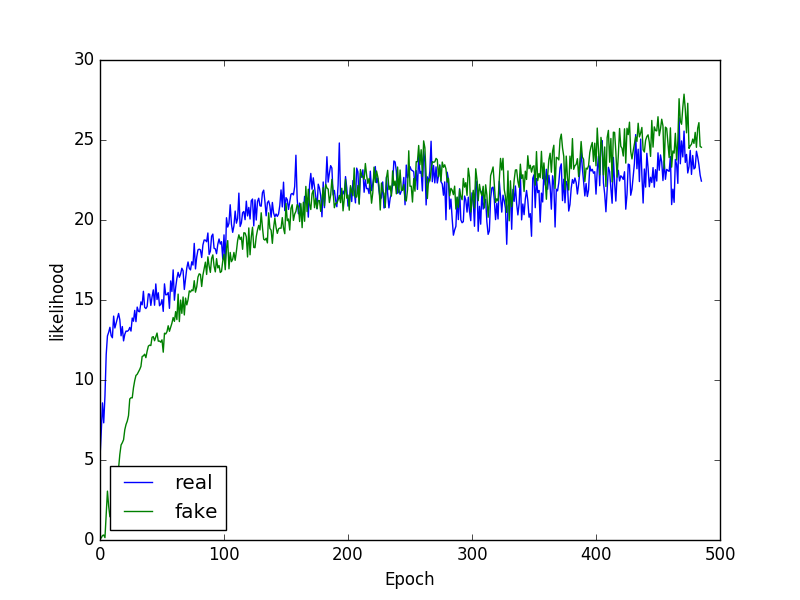}}
\subfigure[$\ell_\mathit{diff}$.]{\label{fig:ap:vanilla_gan_diff}\includegraphics[width=0.24\columnwidth]{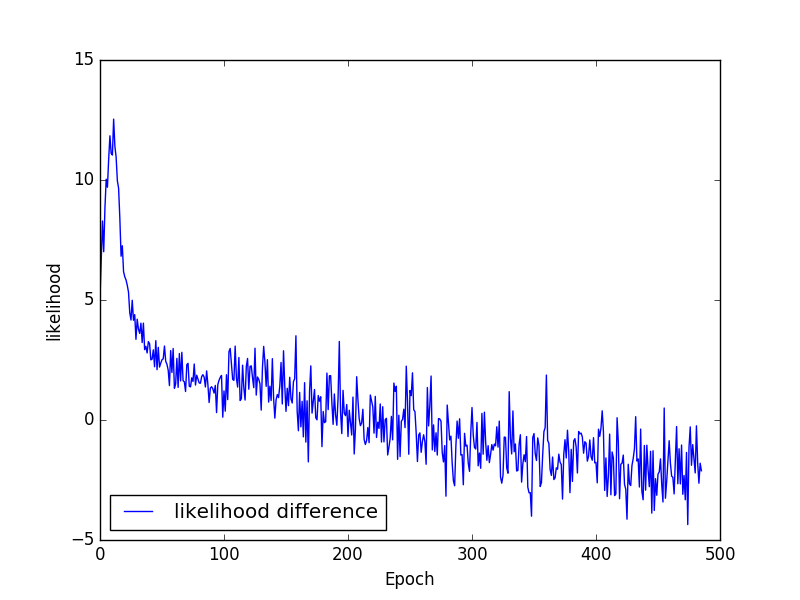}}
\subfigure[$\ell_\mathit{ratio}$.]{\label{fig:ap:vanilla_gan_ratio}\includegraphics[width=0.24\columnwidth]{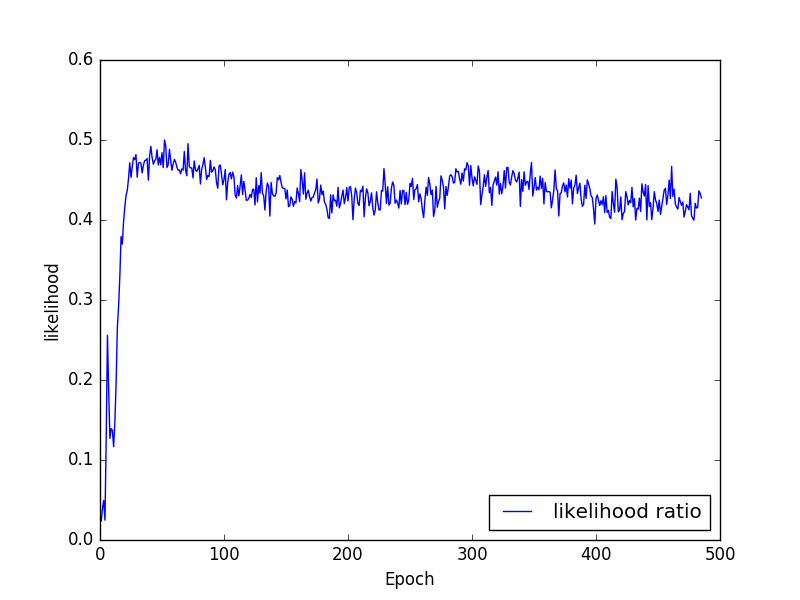}}
\subfigure[Embeddings histogram of the discriminator at\newline epoch 50.]{\label{fig:w_gan_distance}\includegraphics[width=0.24\columnwidth]
{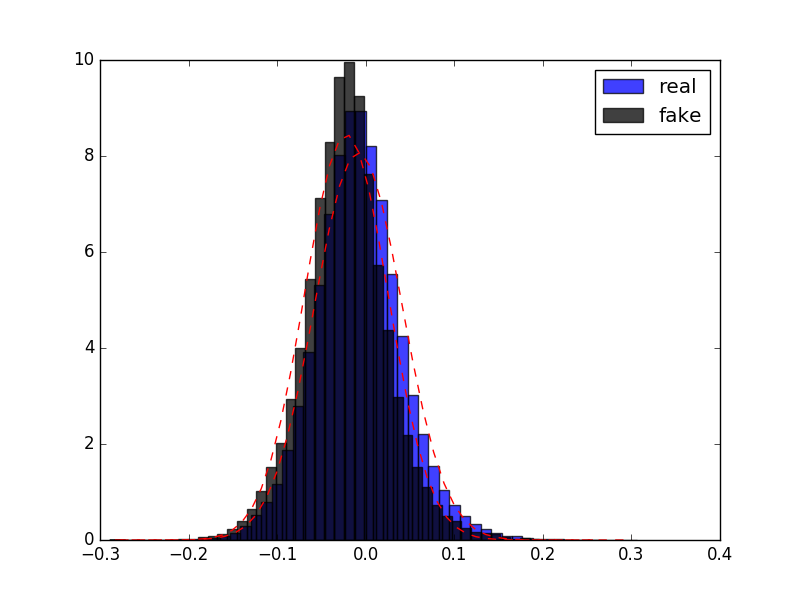}}
\subfigure[Embeddings histogram of the discriminator at\newline epoch 490.]{\label{fig:w_gan_distance}\includegraphics[width=0.24\columnwidth]
{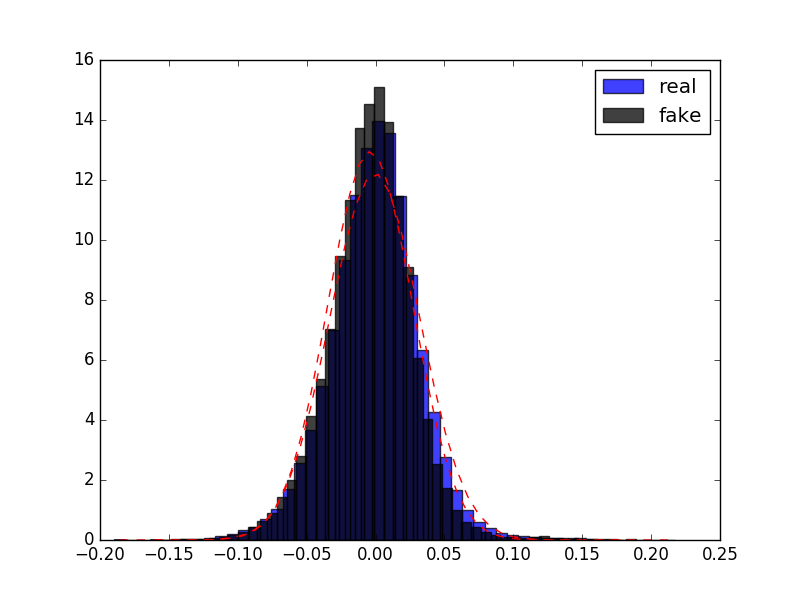}}
\caption{First row: Losses and likelihoods for Discriminator and Generator, LeGAN measures for the Wasserstein GAN \citep{arjovsky2017wasserstein} with weight-clipping. Second row: histogram of the discriminator embedding.
}
\vskip -0.2in
\label{fig:ap:wasserstein_legan}
\end{figure}

In Figure \ref{fig:ap:wasserstein_legan}, we compare the LeGAN measures with the Wasserstein distance in a Wasserstein GAN.
As can be seen, our measures also correlate with the Wasserstein distance as the quality of the generated images improve.

\begin{figure}
\vskip 0.2in
\centering     
\subfigure[Epoch 1.]{\label{fig:0_gpgan_cifar10_embed}\includegraphics[width=0.33\columnwidth]{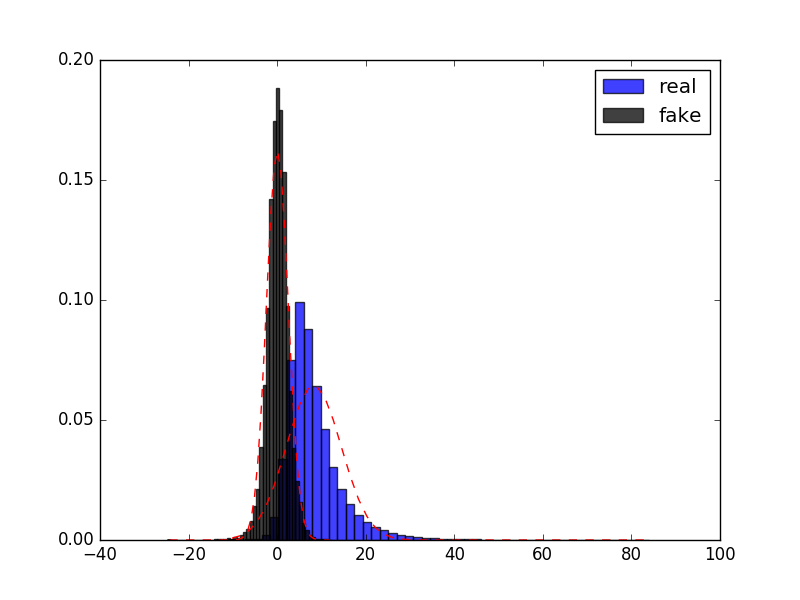}}
\subfigure[Epoch 10.]{\label{fig:10_gpgan_cifar10_embed}\includegraphics[width=0.33\columnwidth]{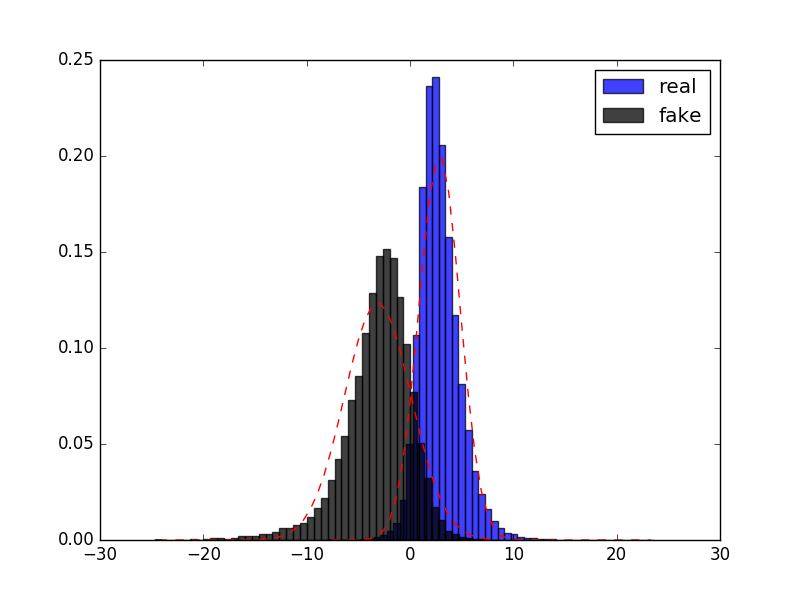}}
\subfigure[Epoch 20.]{\label{fig:20_gpgan_cifar10_embed}\includegraphics[width=0.33\columnwidth]{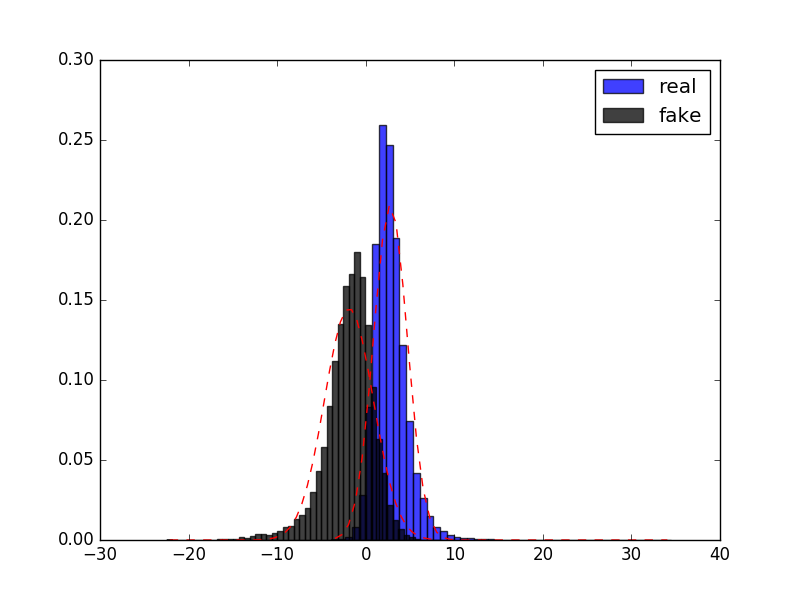}}
\subfigure[Epoch 30.]{\label{fig:30_gpgan_cifar10_embed}\includegraphics[width=0.33\columnwidth]{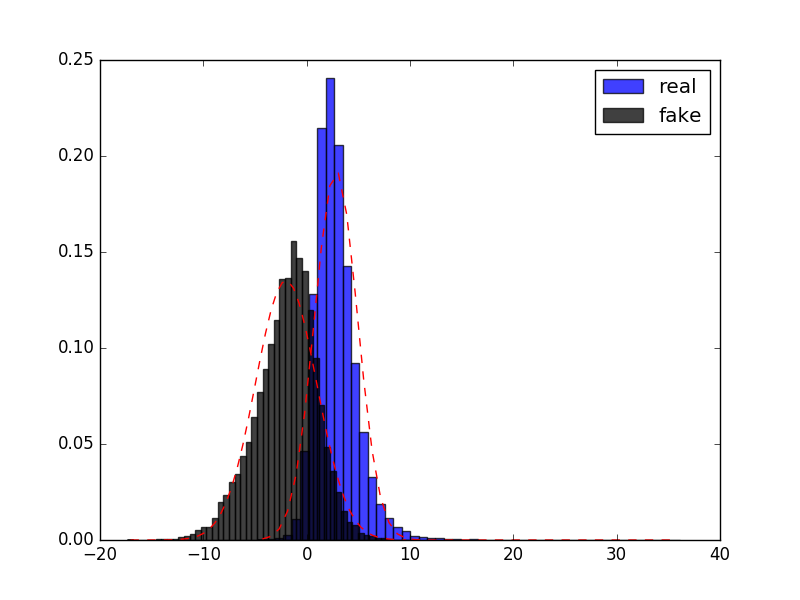}}
\subfigure[Epoch 40.]{\label{fig:40_gpgan_cifar10_embed}\includegraphics[width=0.33\columnwidth]{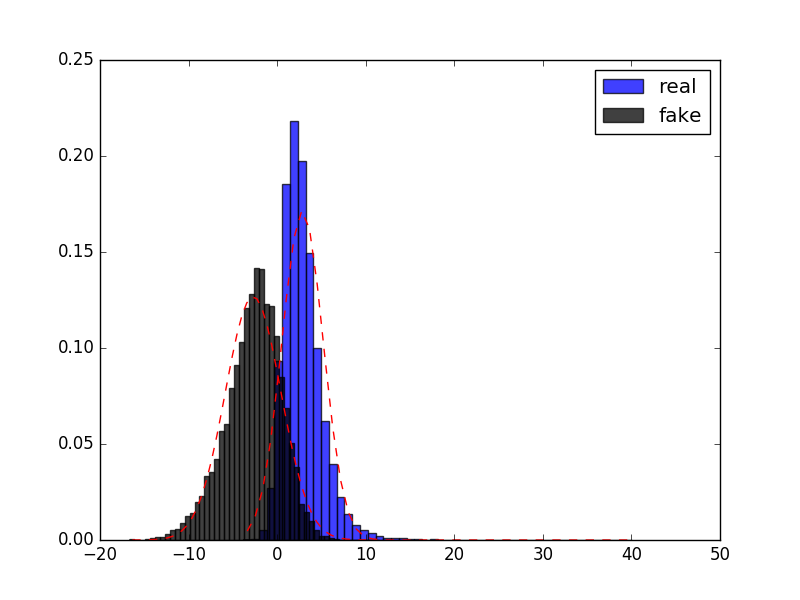}}
\subfigure[Epoch 50.]{\label{fig:wasserstein_gan_distance}\includegraphics[width=0.33\columnwidth]{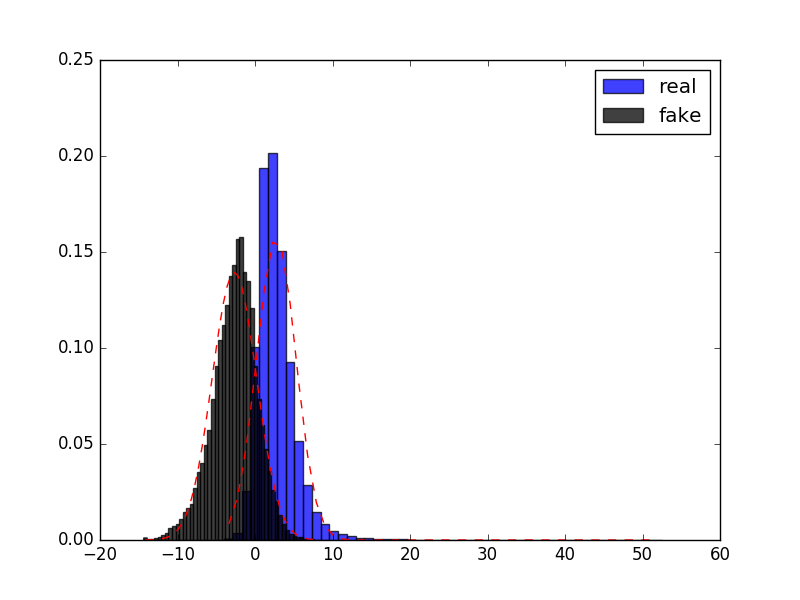}}
\caption{Histograms of the embeddings in different epochs for the vanilla GAN \citep{goodfellow2014generative} without weight-clipping.}
\vskip -0.2in
\label{fig:ap:embedding_dist_nwc}
\end{figure}

In Figure \ref{fig:ap:embedding_dist_nwc}, we provide more examples from the histogram of the discriminator's embeddings for real and fake images without using the weight-clipping.

\begin{figure}
\vskip 0.2in
\centering     
\subfigure[Epoch 1.]{\label{fig:0_gpgan_cifar10_embed}\includegraphics[width=0.33\columnwidth]{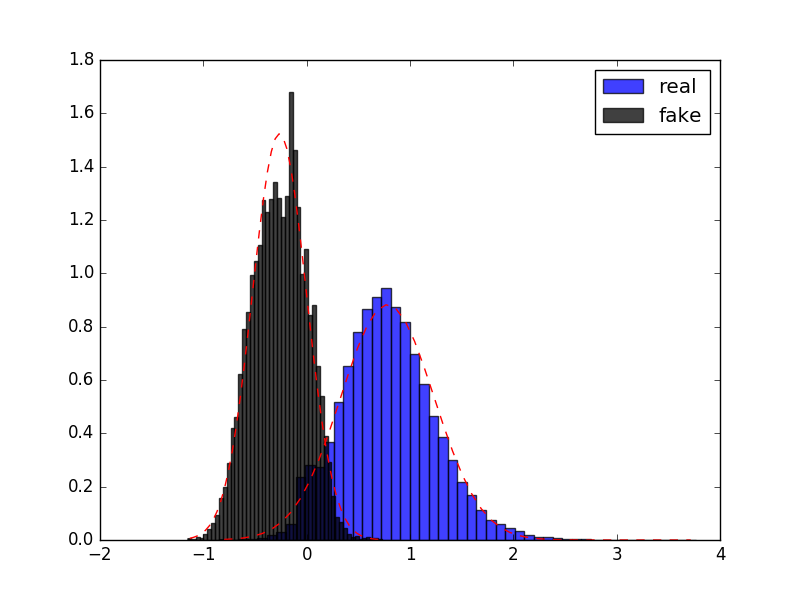}}
\subfigure[Epoch 10.]{\label{fig:10_gpgan_cifar10_embed}\includegraphics[width=0.33\columnwidth]{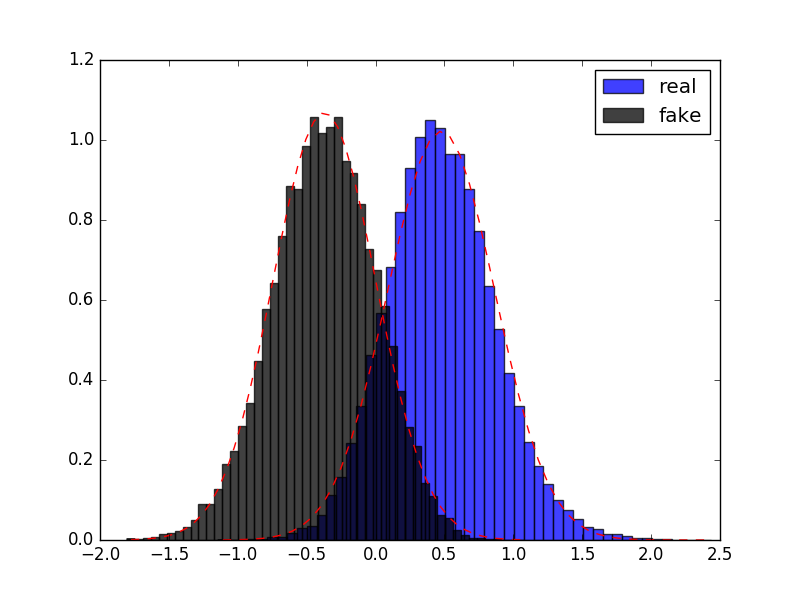}}
\subfigure[Epoch 20.]{\label{fig:20_gpgan_cifar10_embed}\includegraphics[width=0.33\columnwidth]{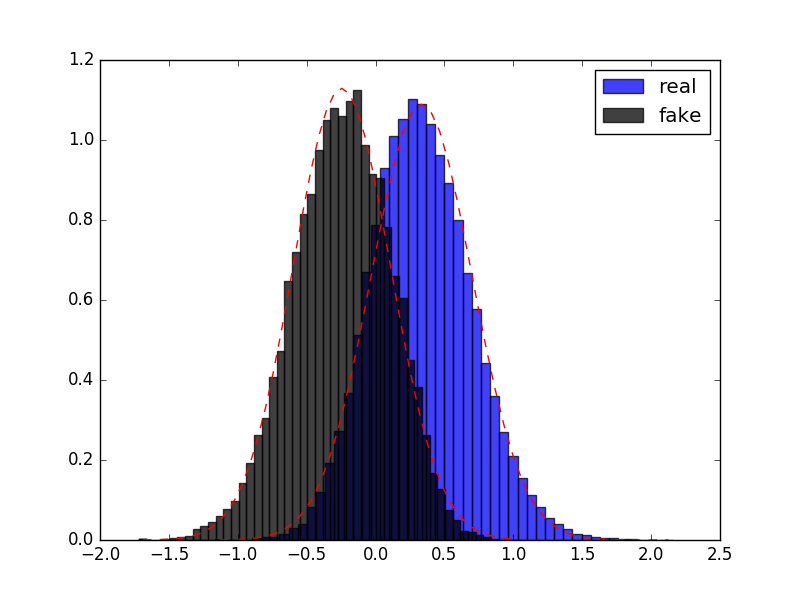}}
\subfigure[Epoch 30.]{\label{fig:30_gpgan_cifar10_embed}\includegraphics[width=0.33\columnwidth]{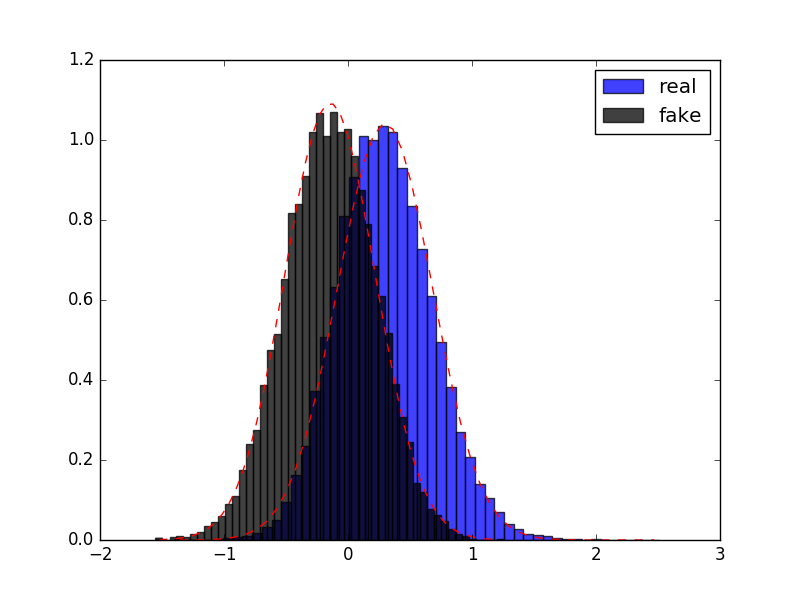}}
\subfigure[Epoch 40.]{\label{fig:40_gpgan_cifar10_embed}\includegraphics[width=0.33\columnwidth]{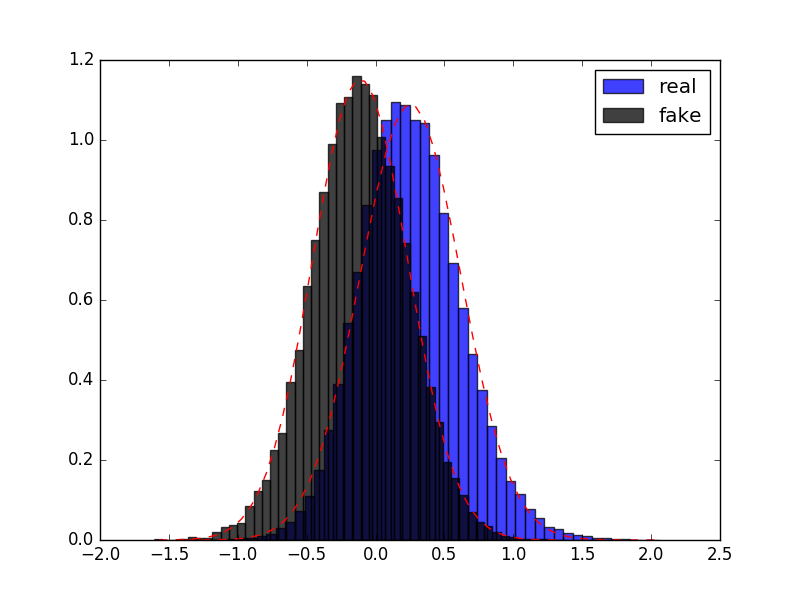}}
\subfigure[Epoch 50.]{\label{fig:wasserstein_gan_distance}\includegraphics[width=0.33\columnwidth]{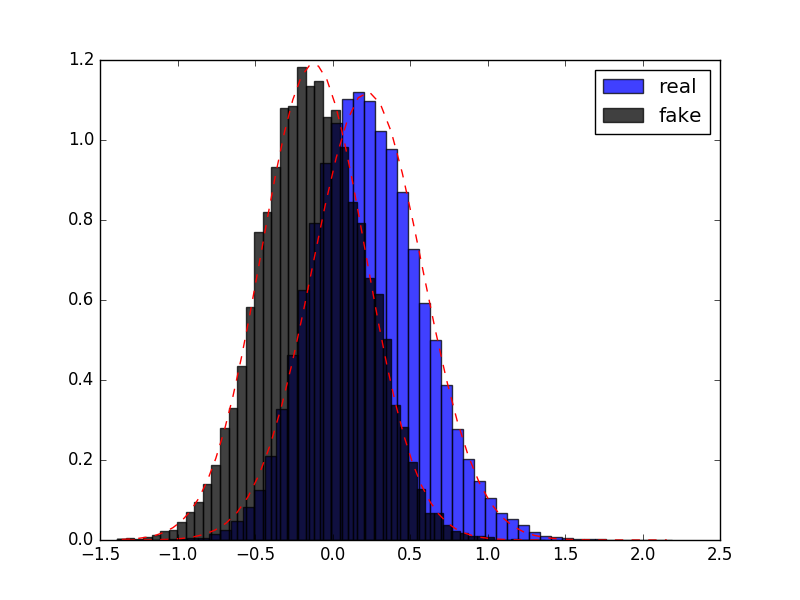}}
\caption{Histograms of the embeddings in different epochs for the vanilla GAN \citep{goodfellow2014generative} with weight-clipping.}
\vskip -0.2in
\label{fig:ap:embedding_dist_wc}
\end{figure}

Figure \ref{fig:ap:embedding_dist_wc} shows more examples from the histogram of the discriminator's embeddings when weight-clipping is applied as discussed in Section \ref{sec:weight_clipping}.

\begin{figure}
\vskip 0.2in
\centering     
\subfigure[${\mathcal{L}_{D}}$ and ${\mathcal{L}_{G}}$.]{\label{fig:ap:vanilla_gan_loss}\includegraphics[width=0.24\columnwidth]{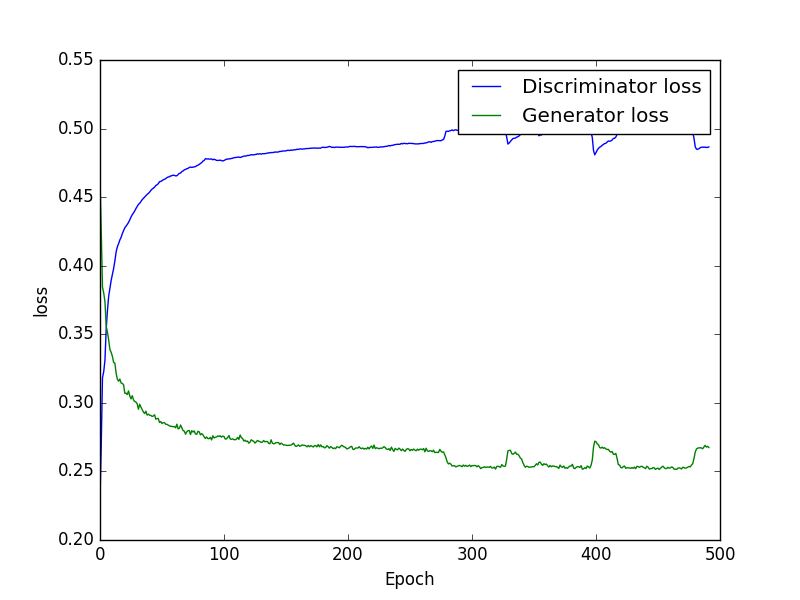}}
\subfigure[$\ell_\mathit{real}$ and $\ell_\mathit{fake}$.]{\label{fig:ap:vanilla_gan_lks}\includegraphics[width=0.24\columnwidth]{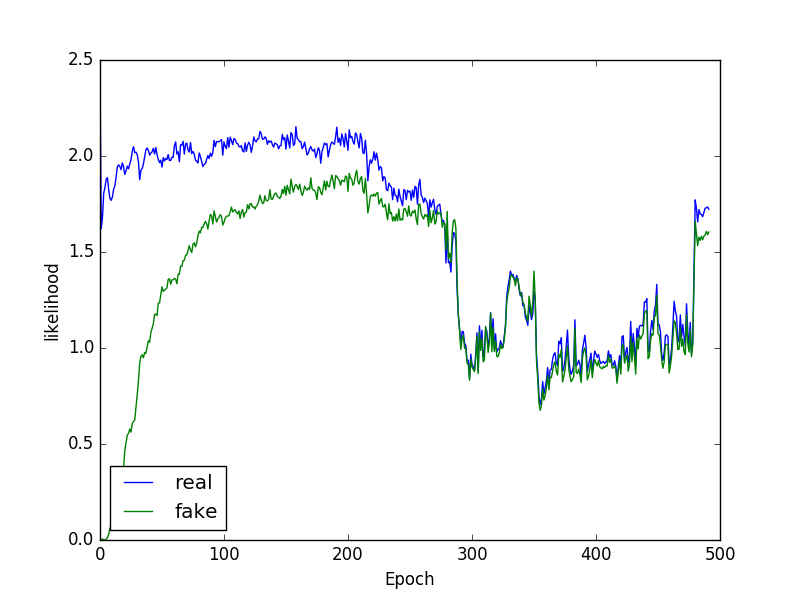}}
\subfigure[$\ell_\mathit{diff}$.]{\label{fig:ap:vanilla_gan_diff}\includegraphics[width=0.24\columnwidth]{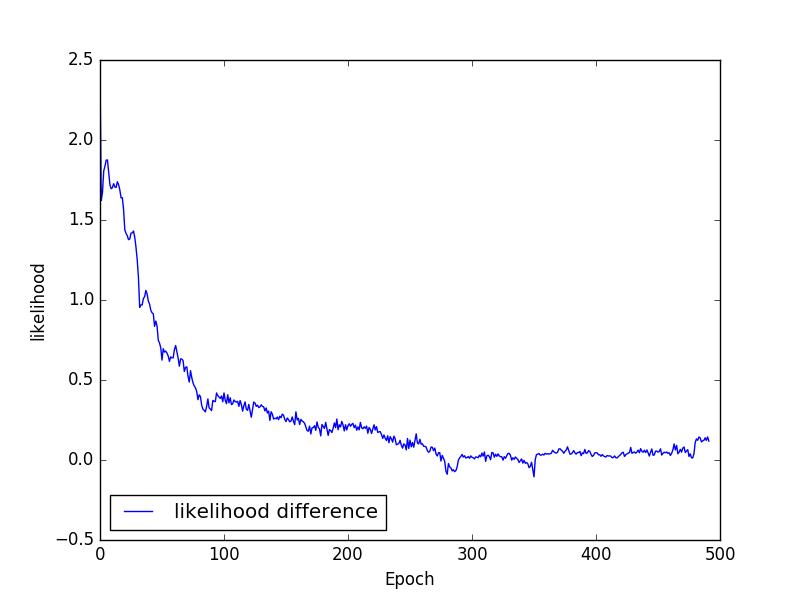}}
\subfigure[$\ell_\mathit{ratio}$.]{\label{fig:ap:vanilla_gan_ratio}\includegraphics[width=0.24\columnwidth]{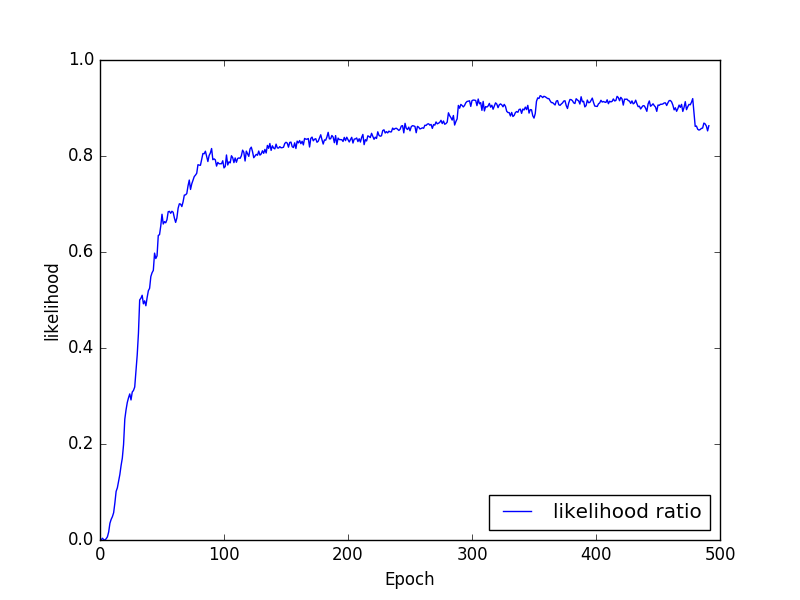}}
\subfigure[Embeddings histogram of the discriminator at\newline epoch 50.]{\label{fig:w_gan_distance}\includegraphics[width=0.24\columnwidth]
{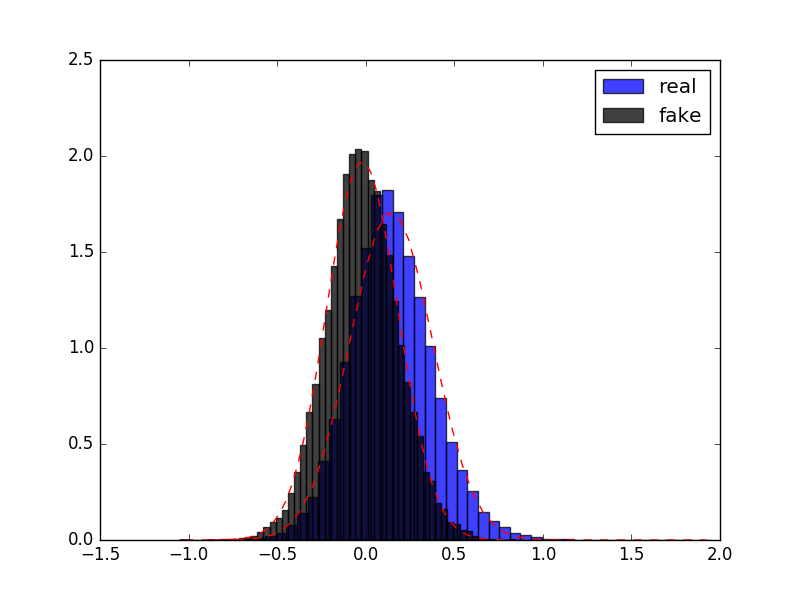}}
\subfigure[Embeddings histogram of the discriminator at\newline epoch 490.]{\label{fig:w_gan_distance}\includegraphics[width=0.24\columnwidth]
{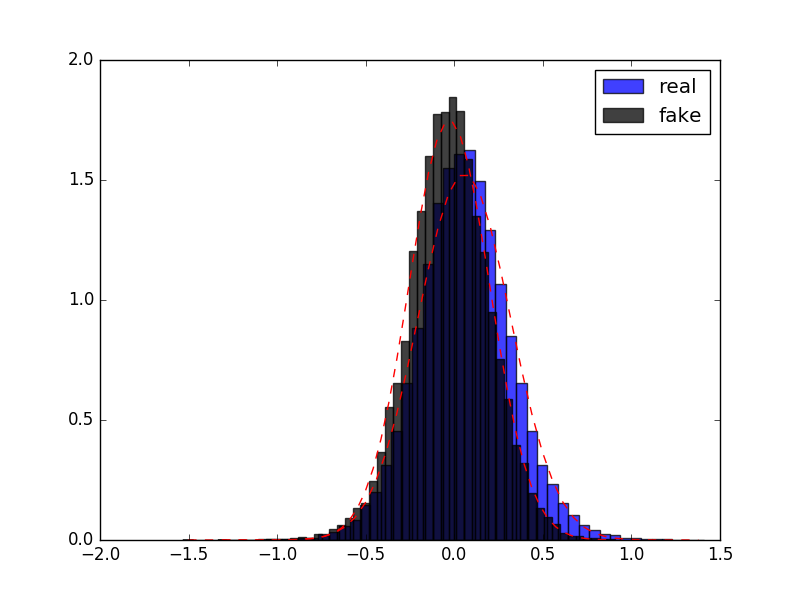}}
\caption{First row: Losses and likelihoods for Discriminator and Generator, LeGAN measures for the LSGAN \citep{mao2016least} with weight-clipping. 
Second row: histogram of the discriminator embedding.
}
\vskip -0.2in
\label{fig:ap:ls_legan}
\end{figure}

\begin{figure}
\vskip 0.2in
\centering     
\subfigure[Vanilla GAN \citep{goodfellow2014generative}.]{\label{fig:ap:vanilla_gan_distance}\includegraphics[width=0.33\columnwidth]{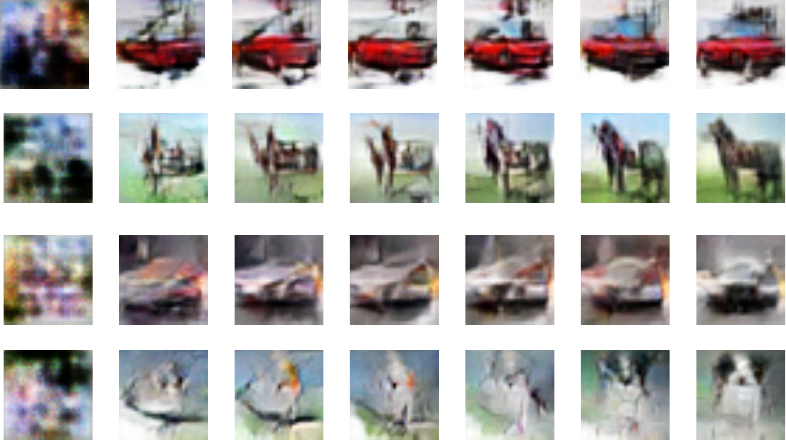}}
\subfigure[LSGAN \citep{mao2016least}.]{\label{fig:w_gan_distance}\includegraphics[width=0.33\columnwidth]
{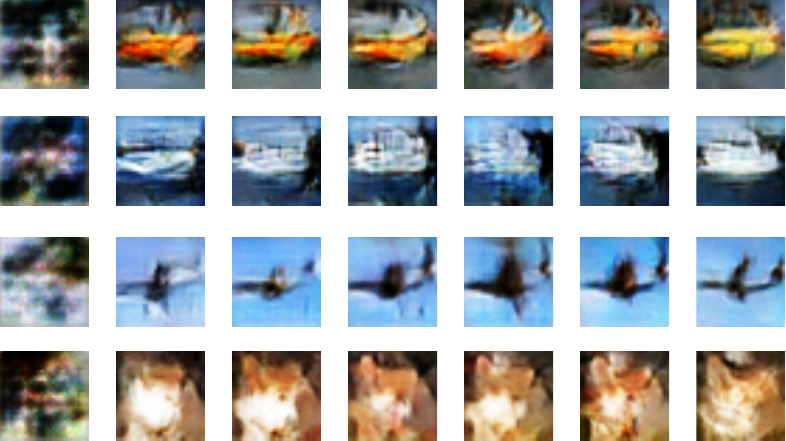}}
\subfigure[Wasserstein GAN \citep{arjovsky2017wasserstein}.]{\label{fig:w_gan_distance}\includegraphics[width=0.33\columnwidth]
{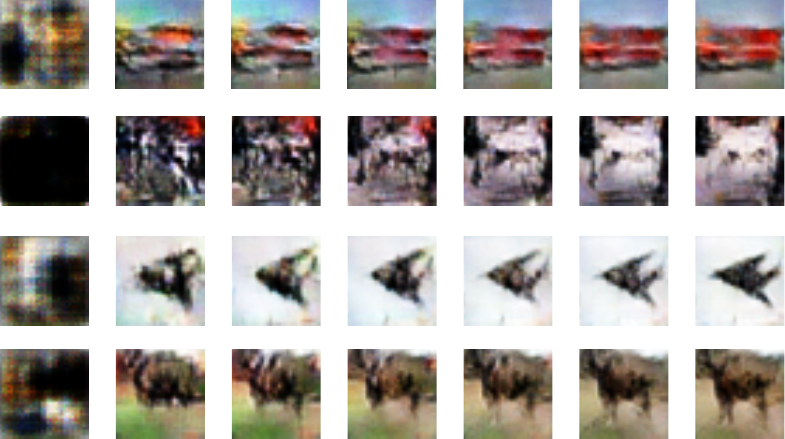}}
\caption{More examples of generated images.}
\vskip -0.2in
\label{fig:ap:gan_evol_examples}
\end{figure}

More examples of LeGAN measures for a vanilla GAN can be found in Figure \ref{fig:ap:gan_evol_examples}.
Also in Figure \ref{fig:ap:ls_legan}, we provide more examples of LeGAN measures with samples of generated images for LSGAN \citep{mao2016least}.


\begin{figure}
\vskip 0.2in
\centering     
\subfigure[${\mathcal{L}_{D}}$ and ${\mathcal{L}_{G}}$.]{\label{fig:ap:vanilla_gan_loss}\includegraphics[width=0.24\columnwidth]{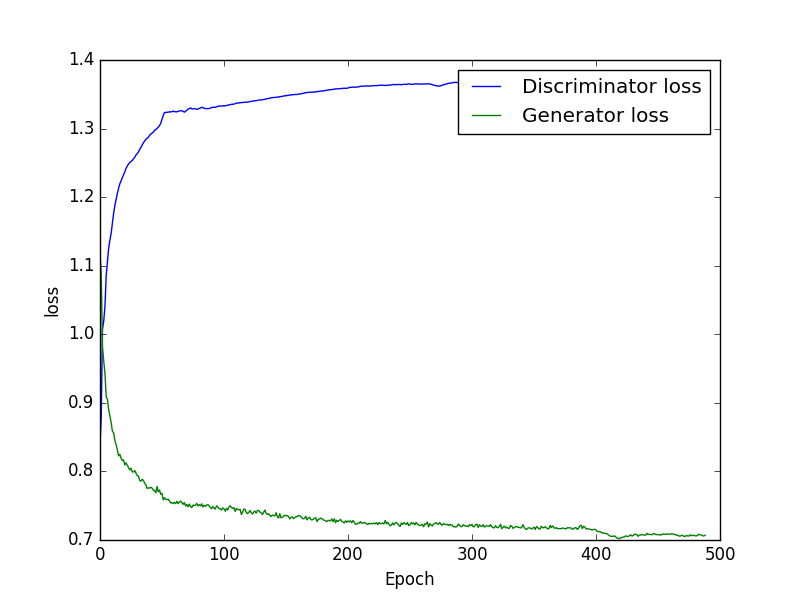}}
\subfigure[$\ell_\mathit{real}$ and $\ell_\mathit{fake}$.]{\label{fig:ap:vanilla_gan_lks}\includegraphics[width=0.24\columnwidth]{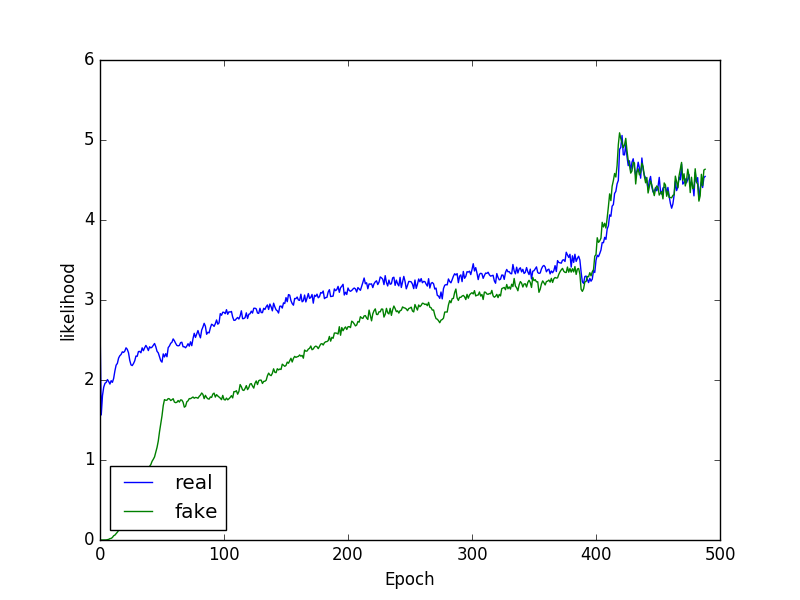}}
\subfigure[$\ell_\mathit{diff}$.]{\label{fig:ap:vanilla_gan_diff}\includegraphics[width=0.24\columnwidth]{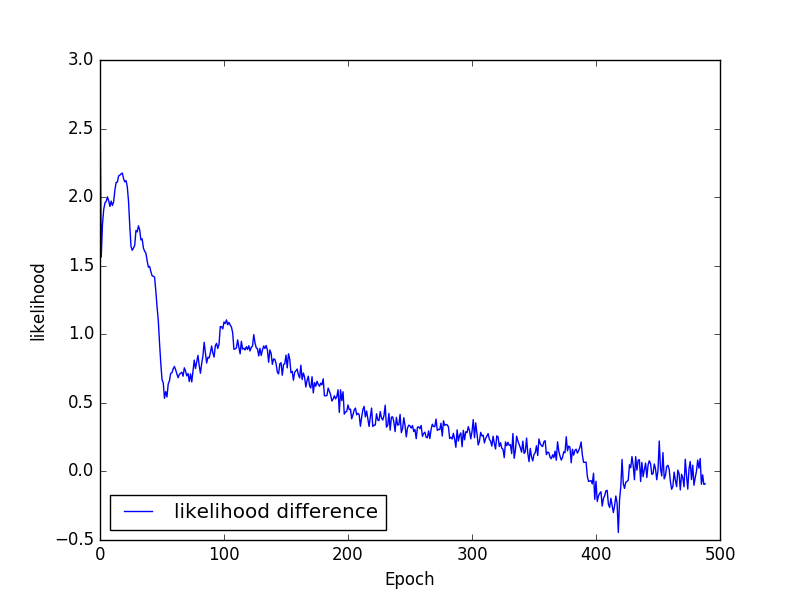}}
\subfigure[$\ell_\mathit{ratio}$.]{\label{fig:ap:vanilla_gan_ratio}\includegraphics[width=0.24\columnwidth]{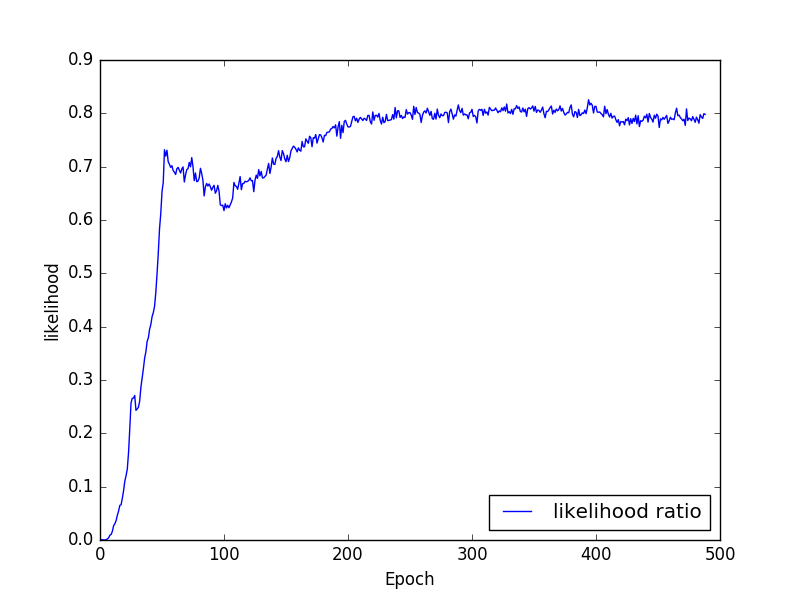}}
\subfigure[Embeddings histogram of the discriminator at\newline epoch 50.]{\label{fig:w_gan_distance}\includegraphics[width=0.24\columnwidth]
{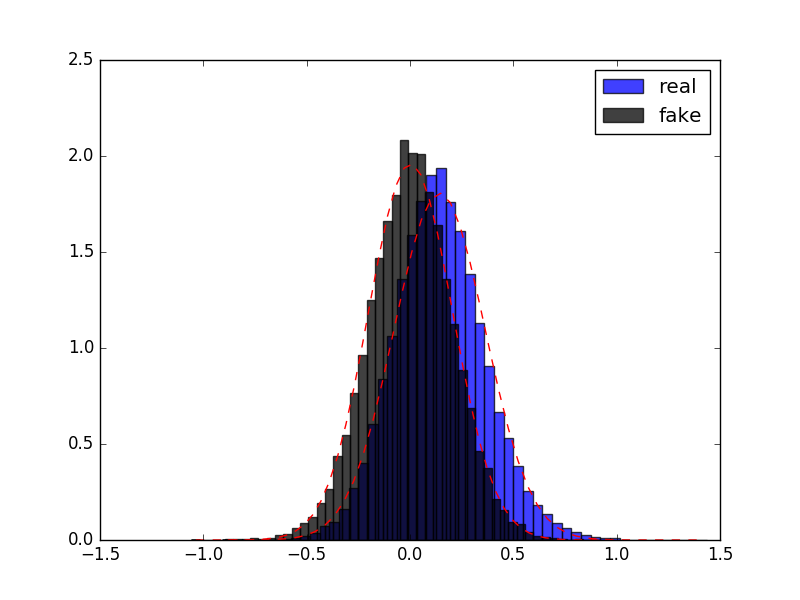}}
\subfigure[Embeddings histogram of the discriminator at\newline epoch 490.]{\label{fig:w_gan_distance}\includegraphics[width=0.24\columnwidth]
{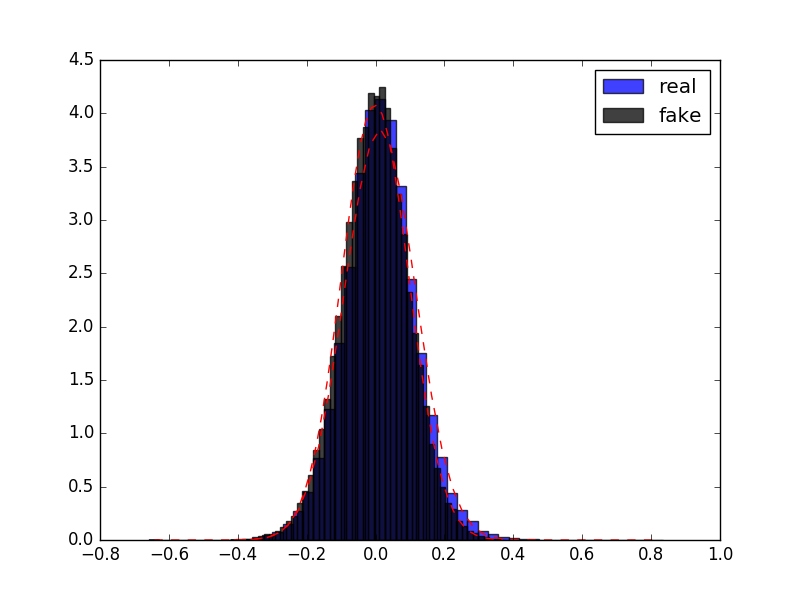}}
\caption{First row: Losses and likelihoods for Discriminator and Generator, LeGAN measures for the vanilla GAN \citep{goodfellow2014generative} with weight-clipping.
Second row: histogram of the discriminator embedding.
}
\vskip -0.2in
\label{fig:ap:vanilla_legan}
\end{figure}

\subsection{Architectures}

Our neural networks are implemented in Python using the \emph{Lasagne library} \citep{lasagne} which is based on \emph{Theano} \citep{bastien2012theano}.

We observed that applying the weight-clipping reduced the power of our discriminator, therefore we updated our discriminator more often (5 times more than the generator in each epoch).

We used the batch size of 128 and the learning rate of $0.0001$ with \emph{ADAM} \citep{kingma2014adam} optimizer. We also used RMSProp and achieved similar results. Therefore we only report the results using ADAM.

In Table \ref{tab:ap:gen_arch}, you can find the architecture of the generator and Table \ref{tab:ap:disc_arch} provides the architecture of the discriminator.

Conv: Convolutional layer. BN: Batch-normalization layer \citep{ioffe2015batch}. LReLU: Leaky rectified activation function \citep{maas2013rectifier}.
\begin{table*}[t]
\caption{The architecture of the Generator.}
\label{tab:ap:gen_arch}
\vskip 0.15in
\begin{center}
\begin{small}
\begin{sc}
\begin{tabular}{c}
\hline
\abovespace\belowspace
Input $1 \times 100$ \\
TransConv Layer (256, $4 \times 4$, stride=1,crop=0,nonlin=LReLu(0.2),init=Normal(0.02, 0))+BN
\\
TransConv Layer (128, $4 \times 4$, stride=2, crop=1,nonlin=LReLu(0.2),init=Normal(0.02, 0))+BN
\\
TransConv Layer (64, $4 \times 4$, stride=2, crop=1,nonlin=LReLu(0.2),init=Normal(0.02, 0))+BN
\\
TransConv Layer (3, $4 \times 4$, stride=2, crop=1,nonlin=Sigmoid,init=Normal(0.02, 0))+BN
\\
\hline
\hline
\end{tabular}
\end{sc}
\end{small}
\end{center}
\vskip -0.1in
\end{table*}

\begin{table*}[t]
\caption{The architecture of the Discriminator.}
\label{tab:ap:disc_arch}
\vskip 0.15in
\begin{center}
\begin{small}
\begin{sc}
\begin{tabular}{c}
\hline
\abovespace\belowspace
Input $3 \times 32 \times 32$\\
Conv Layer (64, $4 \times 4$, stride=2, pad=1,nonlin=LReLu(0.2),init=Normal(0.02, 0))
\\
Conv Layer (128, $4 \times 4$, stride=2, pad=1,LReLu(0.2),init=Normal(0.02, 0))+BN
\\
Conv Layer (256, $2 \times 2$, stride=2, pad=1,LReLu(0.2),init=Normal(0.02, 0))+BN
\\
Conv Layer (1, $4 \times 4$, stride=2, pad=1,LReLu(0.2)\footnote{Except for Wasserstein, which has no nonlinearity. Also, for LeGAN measurements, we use the output of this layer as our embeddings before applying the nonlinearity.},init=Normal(0.02, 0))
\\
\hline
\hline
\end{tabular}
\end{sc}
\end{small}
\end{center}
\vskip -0.1in
\end{table*}

\end{document}


\twocolumn[
\icmltitle{Appendix}
]
\vskip 0.3in

\section{Extended Empirical Results}
\label{extended_results}

\begin{figure*}
\vskip 0.2in
\centering     
\subfigure[$\ell_\mathit{diff}$.]{\label{fig:wasserstein_gan_diff}\includegraphics[width=0.32\textwidth]{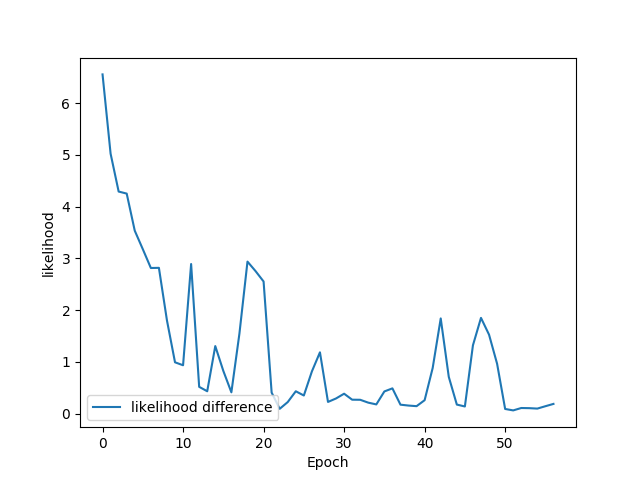}}
\subfigure[$\ell_\mathit{ratio}$.]{\label{fig:wasserstein_gan_ratio}\includegraphics[width=0.32\textwidth]{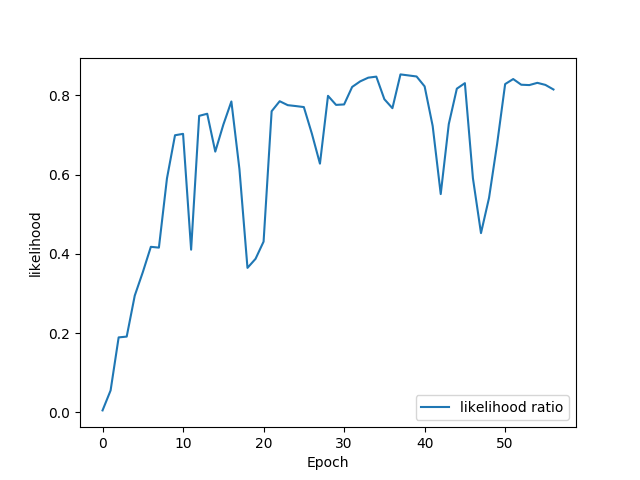}}
\subfigure[Wasserstein distance.]{\label{fig:wasserstein_gan_distance}\includegraphics[width=0.33\textwidth]{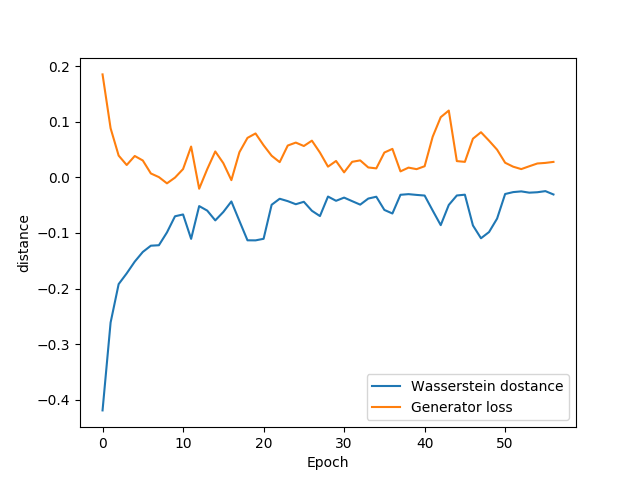}}
\hfill
\caption{Different measures for a Wasserstein GAN \citep{arjovsky2017wasserstein}.}
\vskip -0.2in
\label{fig:wasserstein_gan}
\end{figure*}

In Figure \ref{fig:wasserstein_gan}, we compare the LeGAN measures with the Wasserstein distance in a Wasserstein GAN.
As can be seen, our measures also correlate with the Wasserstein distance as the quality of the generated images improve.

\begin{figure}
\vskip 0.2in
\centering     
\subfigure[Epoch 1.]{\label{fig:0_gpgan_cifar10_embed}\includegraphics[width=0.49\columnwidth]{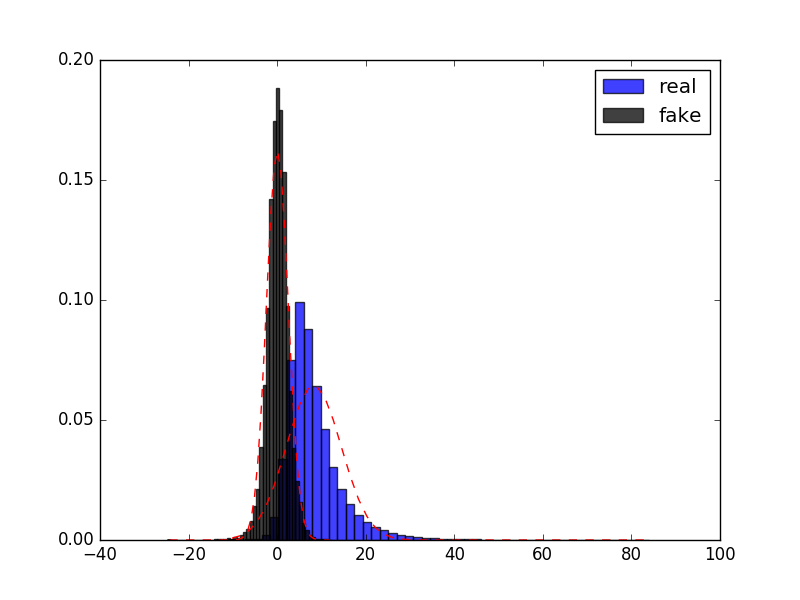}}
\subfigure[Epoch 10.]{\label{fig:10_gpgan_cifar10_embed}\includegraphics[width=0.49\columnwidth]{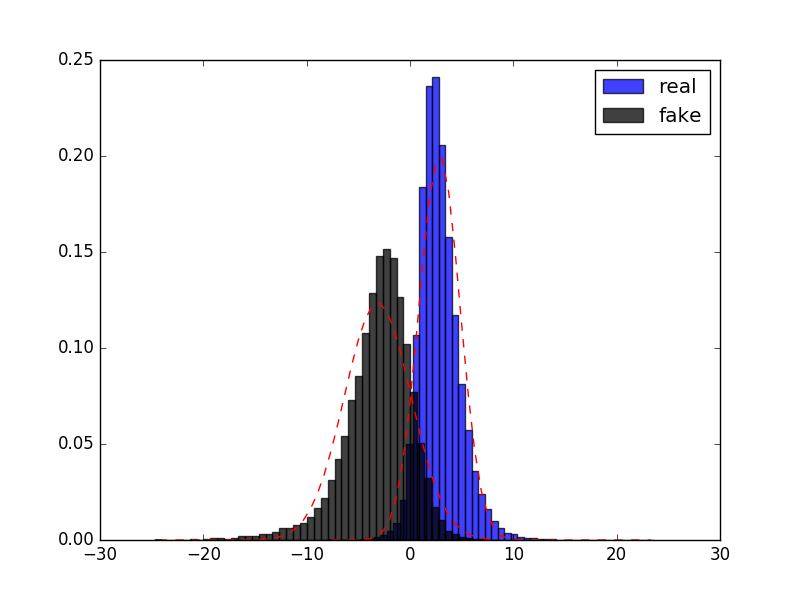}}
\subfigure[Epoch 20.]{\label{fig:20_gpgan_cifar10_embed}\includegraphics[width=0.49\columnwidth]{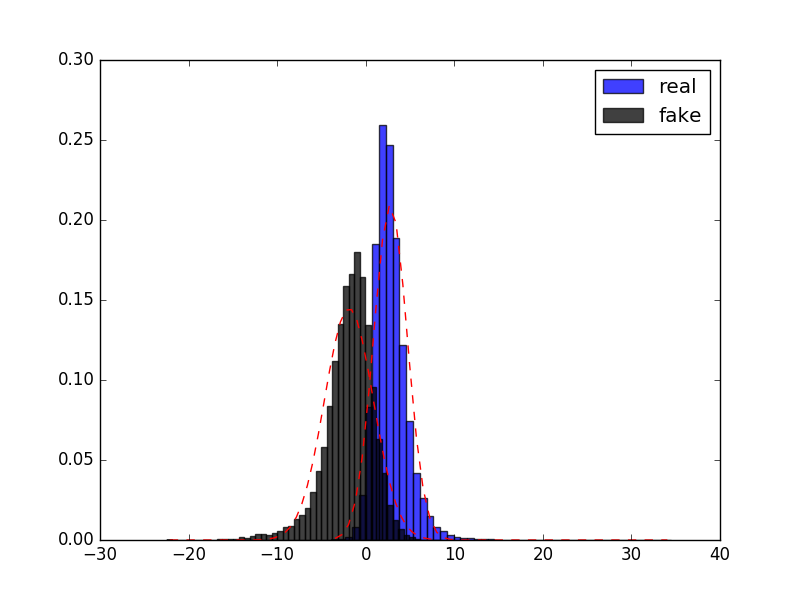}}
\subfigure[Epoch 30.]{\label{fig:30_gpgan_cifar10_embed}\includegraphics[width=0.49\columnwidth]{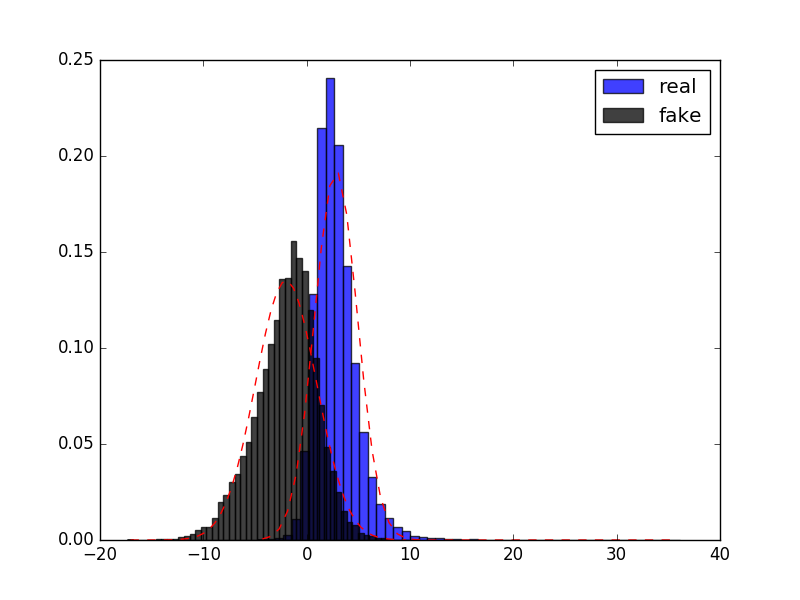}}
\subfigure[Epoch 40.]{\label{fig:40_gpgan_cifar10_embed}\includegraphics[width=0.49\columnwidth]{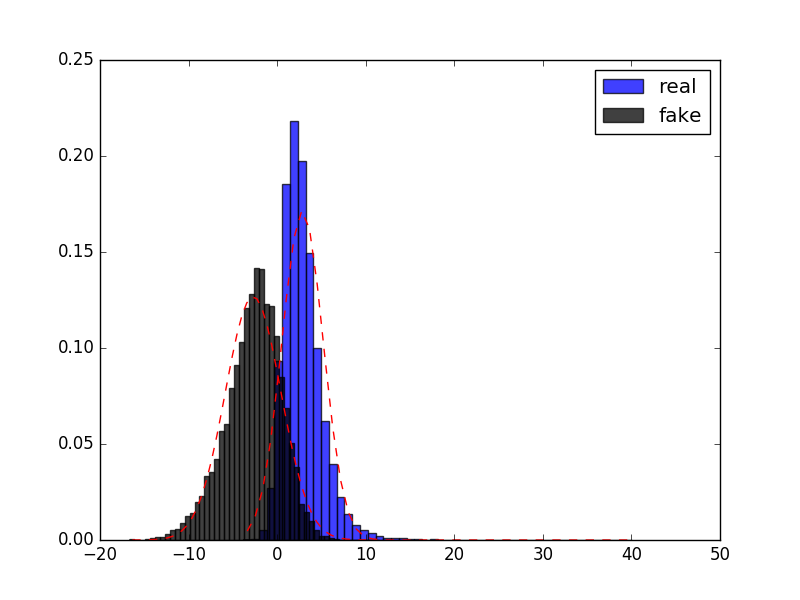}}
\subfigure[Epoch 50.]{\label{fig:wasserstein_gan_distance}\includegraphics[width=0.49\columnwidth]{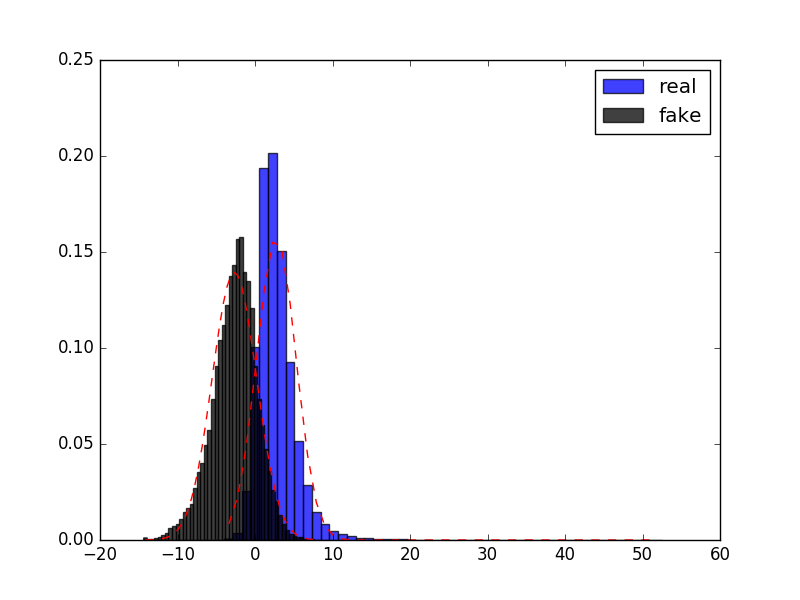}}
\caption{Histograms of the embeddings in different epochs for the vanilla GAN \citep{goodfellow2014generative} without weight-clipping.}
\vskip -0.2in
\label{fig:embedding_dist_nwc}
\end{figure}

In Figure \ref{fig:embedding_dist_nwc}, we provide more examples from the histogram of the discriminator's embeddings for real and fake images without using the weight-clipping.

\begin{figure}
\vskip 0.2in
\centering     
\subfigure[Epoch 1.]{\label{fig:0_gpgan_cifar10_embed}\includegraphics[width=0.49\columnwidth]{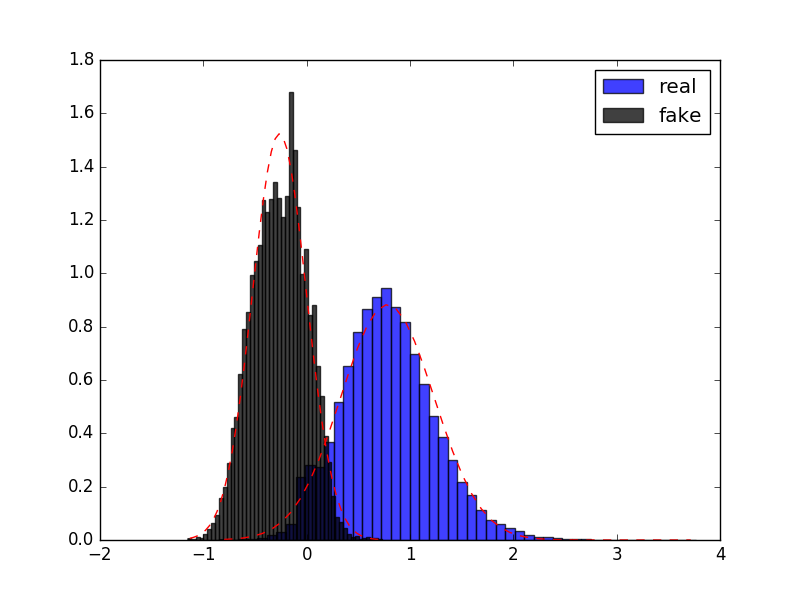}}
\subfigure[Epoch 10.]{\label{fig:10_gpgan_cifar10_embed}\includegraphics[width=0.49\columnwidth]{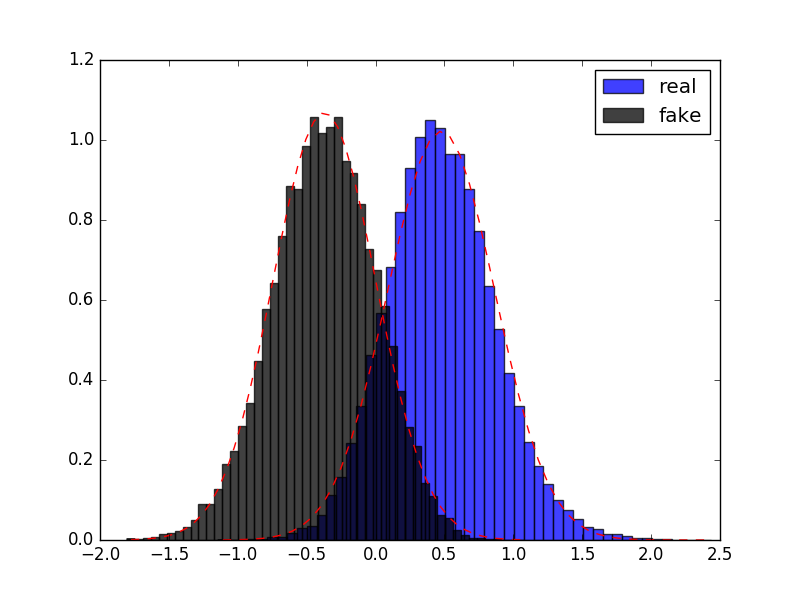}}
\subfigure[Epoch 20.]{\label{fig:20_gpgan_cifar10_embed}\includegraphics[width=0.49\columnwidth]{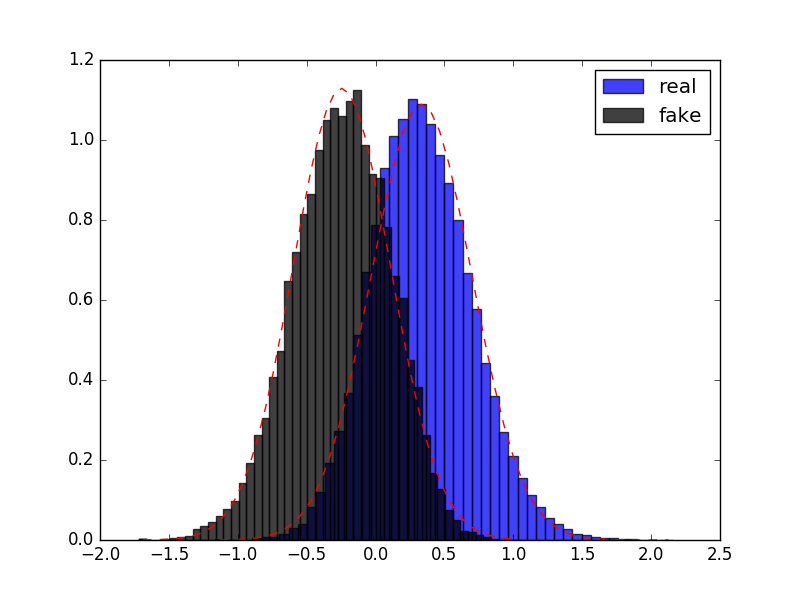}}
\subfigure[Epoch 30.]{\label{fig:30_gpgan_cifar10_embed}\includegraphics[width=0.49\columnwidth]{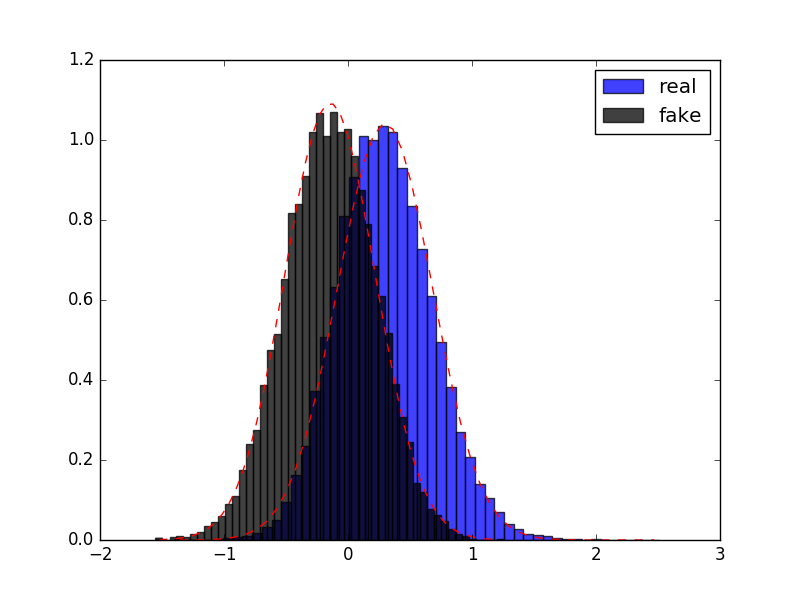}}
\subfigure[Epoch 40.]{\label{fig:40_gpgan_cifar10_embed}\includegraphics[width=0.49\columnwidth]{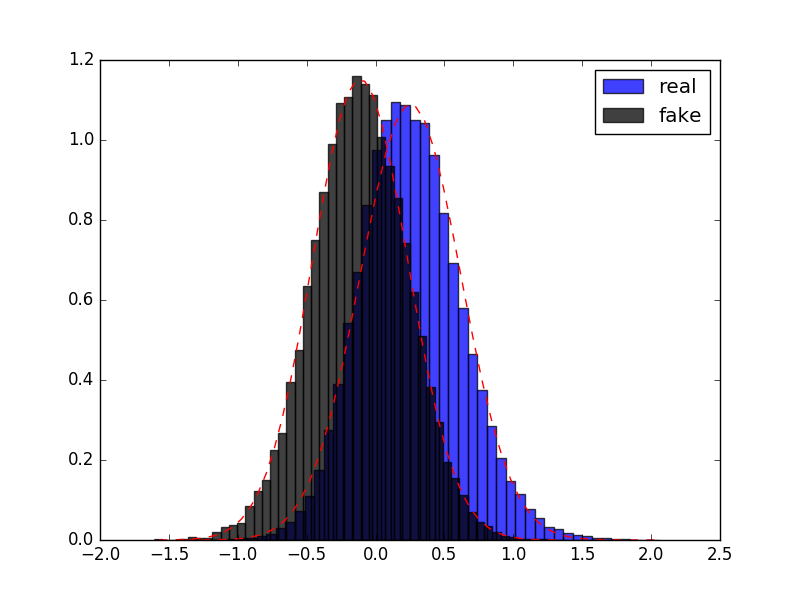}}
\subfigure[Epoch 50.]{\label{fig:wasserstein_gan_distance}\includegraphics[width=0.49\columnwidth]{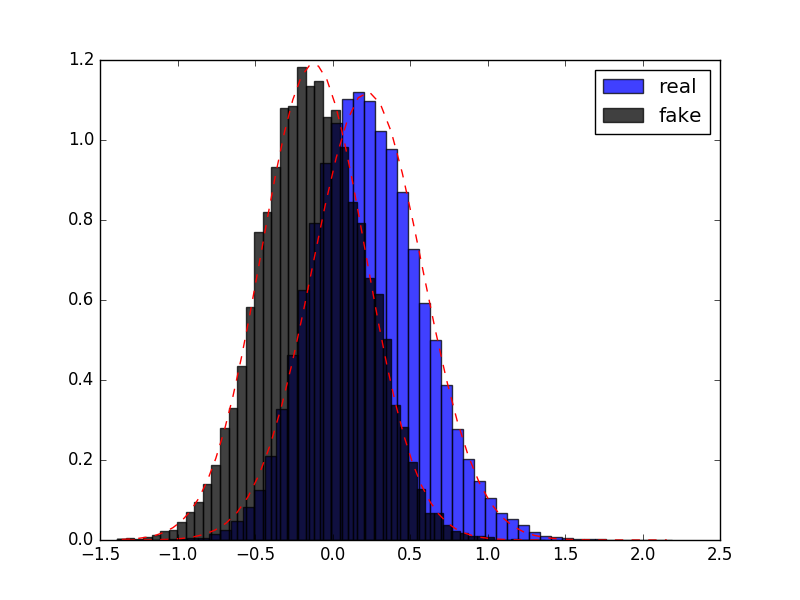}}
\caption{Histograms of the embeddings in different epochs for the vanilla GAN \citep{goodfellow2014generative} with weight-clipping.}
\vskip -0.2in
\label{fig:embedding_dist_wc}
\end{figure}

Figure \ref{fig:embedding_dist_wc} shows more examples from the histogram of the discriminator's embeddings when weight-clipping is applied as discussed in Section 4.

\begin{figure}
\vskip 0.2in
\centering     
\subfigure[$\ell_\mathit{diff}$.]{\label{fig:wasserstein_gan_diff}\includegraphics[width=0.49\columnwidth]{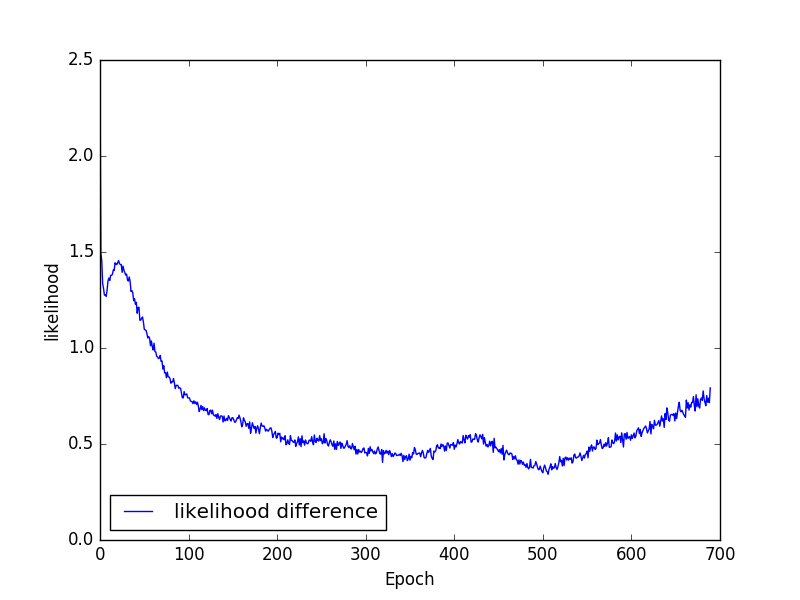}}
\subfigure[$\ell_\mathit{ratios}$.]{\label{fig:wasserstein_gan_ratio}\includegraphics[width=0.49\columnwidth]{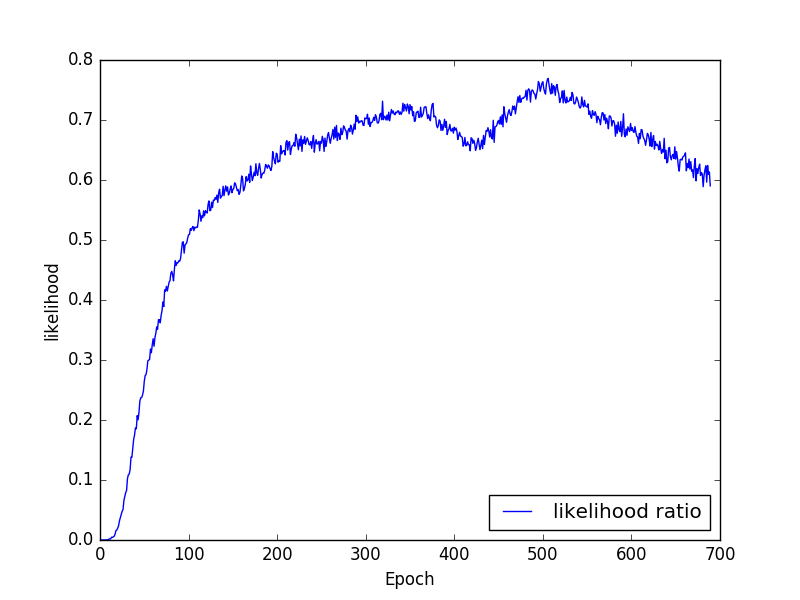}}
\subfigure[Generated images at\newline epoch 500.]{\label{fig:wasserstein_gan_distance}\includegraphics[width=0.49\columnwidth]
{final_ls/500_gpgan_cifar10_fixedsamples}}
\subfigure[Generated images at\newline epoch 690.]{\label{fig:wasserstein_gan_distance}\includegraphics[width=0.49\columnwidth]
{final_ls/690_gpgan_cifar10_fixedsamples}}
\caption{First two rows: LeGAN measures for the LSGAN \citep{mao2016least} with weight-clipping. Third row: examples of generated images.}
\vskip -0.2in
\label{fig:ls_legan}
\end{figure}

More examples of LeGAN measures for a vanilla GAN can be found in Figure \ref{fig:orig_legan}.
Also in Figure \ref{fig:ls_legan}, we provide more examples of LeGAN measures with samples of generated images for LSGAN \citep{mao2016least}.

\begin{figure}
\vskip 0.2in
\centering     
\subfigure[$\ell_\mathit{diff}$.]{\label{fig:wasserstein_gan_diff}\includegraphics[width=0.49\columnwidth]{final_org/gpgan_cifar10_diff}}
\subfigure[$\ell_\mathit{ratio}$.]{\label{fig:wasserstein_gan_ratio}\includegraphics[width=0.49\columnwidth]{final_org/gpgan_cifar10_likelihood_ratios}}
\subfigure[Generated images at\newline epoch 350.]{\label{fig:wasserstein_gan_distance}\includegraphics[width=0.49\columnwidth]{final_org/350_gpgan_cifar10_fixedsamples}}
\subfigure[Generated images at\newline epoch 728.]{\label{fig:wasserstein_gan_distance}\includegraphics[width=0.49\columnwidth]{final_org/728_gpgan_cifar10_fixedsamples}}
\caption{First two rows: LeGAN measures for the vanilla GAN \citep{goodfellow2014generative} with weight-clipping. Third row: examples of generated images.}
\vskip -0.2in
\label{fig:orig_legan}
\end{figure}

\section{Architectures}

Our neural networks are implemented in Python using the \emph{Lasagne library} \citep{lasagne} which is based on \emph{Theano} \citep{bastien2012theano}.

We observed that applying the weight-clipping reduced the power of our discriminator, therefore we updated our discriminator more often (5 times more than the generator in each epoch).

We used the batch size of 128 and the learning rate of $0.0001$ with \emph{ADAM} \citep{kingma2014adam} optimizer. We also used RMSProp and achieved similar results. Therefore we only report the results using ADAM.

In Table \ref{tab:gen_arch}, you can find the architecture of the generator and Table \ref{tab:disc_arch} provides the architecture of the discriminator.

Conv: Convolutional layer. BN: Batch-normalization layer \citep{ioffe2015batch}. LReLU: Leaky rectified activation function \citep{maas2013rectifier}.
\begin{table*}[t]
\caption{The architecture of the Generator.}
\label{tab:gen_arch}
\vskip 0.15in
\begin{center}
\begin{small}
\begin{sc}
\begin{tabular}{c}
\hline
\abovespace\belowspace
Input $1 \times 100$ \\
TransConv Layer (256, $4 \times 4$, stride=1,crop=0,nonlin=LReLu(0.2),init=Normal(0.02, 0))+BN
\\
TransConv Layer (128, $4 \times 4$, stride=2, crop=1,nonlin=LReLu(0.2),init=Normal(0.02, 0))+BN
\\
TransConv Layer (64, $4 \times 4$, stride=2, crop=1,nonlin=LReLu(0.2),init=Normal(0.02, 0))+BN
\\
TransConv Layer (3, $4 \times 4$, stride=2, crop=1,nonlin=Sigmoid,init=Normal(0.02, 0))+BN
\\
\hline
\hline
\end{tabular}
\end{sc}
\end{small}
\end{center}
\vskip -0.1in
\end{table*}

\begin{table*}[t]
\caption{The architecture of the Discriminator.}
\label{tab:disc_arch}
\vskip 0.15in
\begin{center}
\begin{small}
\begin{sc}
\begin{tabular}{c}
\hline
\abovespace\belowspace
Input $3 \times 32 \times 32$\\
Conv Layer (64, $4 \times 4$, stride=2, pad=1,nonlin=LReLu(0.2),init=Normal(0.02, 0))
\\
Conv Layer (128, $4 \times 4$, stride=2, pad=1,LReLu(0.2),init=Normal(0.02, 0))+BN
\\
Conv Layer (256, $2 \times 2$, stride=2, pad=1,LReLu(0.2),init=Normal(0.02, 0))+BN
\\
Conv Layer (1, $4 \times 4$, stride=2, pad=1,LReLu(0.2)\footnote{Except for Wasserstein, which has no nonlinearity. Also, for LeGAN measurements, we use the output of this layer as our embeddings before applying the nonlinearity.},init=Normal(0.02, 0))
\\
\hline
\hline
\end{tabular}
\end{sc}
\end{small}
\end{center}
\vskip -0.1in
\end{table*}

\bibliography{refs}
\bibliographystyle{icml2017}